\documentclass[sigconf]{acmart}
\usepackage{bbm}
\usepackage{subfigure}
\usepackage[dvipdfmx]{}
\usepackage{float}
\usepackage{balance}
\usepackage[english,ngerman]{babel}
\def\BibTeX{{\rm B\kern-.05em{\sc i\kern-.025em b}\kern-.08emT\kern-.1667em\lower.7ex\hbox{E}\kern-.125emX}}
\acmYear{2019} 
\copyrightyear{}
\setcopyright{none}
\acmConference[KDD '19]{The 25th ACM SIGKDD Conference on Knowledge Discovery and Data Mining}{August 4--8, 2019}{Anchorage, AK, USA}
\acmBooktitle{The 25th ACM SIGKDD Conference on Knowledge Discovery and Data Mining (KDD '19), August 4--8, 2019, Anchorage, AK, USA}
\acmPrice{15.00}
\acmDOI{10.1145/3292500.3330937}
\acmISBN{978-1-4503-6201-6/19/08}

\begin{document}
\fancyhead{}
\title{Deep Mixture Point Processes: Spatio-temporal Event Prediction with Rich Contextual Information}

\author{Maya Okawa$^1$, Tomoharu Iwata$^2$, Takeshi Kurashima$^1$, 
Yusuke Tanaka$^1$, Hiroyuki Toda$^1$ and Naonori Ueda$^2$}
\affiliation{\institution{$^1$NTT Service Evolution Labs, NTT Corporation, Kanagawa, Japan}}
\affiliation{\{maya.ookawa.af, takeshi.kurashima.uf, yusuke.tanaka.rh, hiroyuki.toda.xb\}@hco.ntt.co.jp}
\affiliation{\institution{$^2$NTT Communication Science Labs, NTT Corporation, Kyoto, Japan}}
\affiliation{\{tomoharu.iwata.gy, naonori.ueda.fr\}@hco.ntt.co.jp}
\renewcommand{\shortauthors}{}

\begin{abstract}
Predicting when and where events will occur in cities, like taxi pick-ups, crimes, and vehicle collisions, is a challenging and important problem with many applications in fields such as urban planning, transportation optimization and location-based marketing. Though many point processes have been proposed to model events in a continuous spatio-temporal space, none of them allow for the consideration of the rich contextual factors that affect event occurrence, such as weather, social activities, geographical characteristics, and traffic. In this paper, we propose \textsf{DMPP} (Deep Mixture Point Processes), a point process model for predicting spatio-temporal events with the use of rich contextual information; a key advance is its incorporation of the heterogeneous and high-dimensional context available in image and text data. Specifically, we design the intensity of our point process model as a mixture of kernels, where the mixture weights are modeled by a deep neural network. This formulation allows us to automatically learn the complex nonlinear effects of the contextual factors on event occurrence. At the same time, this formulation makes analytical integration over the intensity, which is required for point process estimation, tractable. 
We use real-world data sets from different domains to demonstrate that DMPP has better predictive performance than existing methods. 
\end{abstract}

\maketitle

\section{Introduction}
In cities, large volumes of event data are being generated by human activities. Such event data includes information about time and geolocation, indicating where and when each event occurred.  
For instance, taxi pick-up records are represented as a list of events consisting of the pick-up locations and the departure times. 
Crimes are recorded together with the time and location  
at which the crime took place.  
Predicting events is a key component of applications in many fields
such as urban planning, transportation optimization and location-based marketing. 
If taxi dispatch service operators can estimate with high accuracy the future taxi pick-up times and locations, they can allocate taxis to the right places and the right times in advance.  
Criminal incident prediction will help law enforcement agencies to implement effective police activities that can suppress criminality.  

Predicting spatio-temporal events, however, is extremely challenging, because event occurrence is determined by various contextual factors. 
Such contextual features also include geographical characteristics, e.g., transportation networks \cite{gao2013} and land use \cite{Brantingham1981};  
temporal attributes, e.g., day of week and weather conditions \cite{Zhang2016,Chen2016}; and 
other features, e.g., social and traffic information \cite{Zhou2014,Wang2015,Chen2016}.
This demands a framework that can capture the complex dynamics of event occurrence given the contextual features present. 
The conventional approach to this problem is based on regression models \cite{Zhou2014,Shimosaka2015a,Hoang2016}. 
They are intended to model the aggregated number of events within a predefined spatial region and time interval, 
which is fundamentally different from our task.  
We focus more on the point process approach to model a sequence of events in continuous time and space, without aggregation,   
by using explicit information about location and/or time; and  
predicting the precise time and location at which each event will occur.  

Point process is a sophisticated framework for modeling a sequence of events in continuous time and space; 
it directly estimates an {\sl intensity} function that describes the rate of events occurring at any location and any time.  
The influence of the contextual features can be modeled by special point process models \cite{Cox1992,Hayano2011,Giorgi2003},  
where the intensity function is described as a function of covariates, i.e., the contextual features.  
However, this approach has a fundamental limitation.  
In many practical cases, 
their assumptions on the functional form of covariates may be too restrictive to capture complex and intricate effects of contextual features; 
they do not accommodate unstructured data such as images and texts. 
Most contextual features take the form of unstructured representations.  
For example, information about geographical characteristics can be obtained from map images.
Traffic and social event information can be expressed in the form of natural language expressions.

In this paper, we propose an event prediction method that effectively incorporates such unstructured data into the point process model.  
Motivated by the recent success of the deep learning approach, we use it to enhance the point process model.  
The naive approach is to directly model the intensity by a deep neural network. 
Unfortunately, this approach triggers the intractable optimization problem as integral computations are required to determine the likelihood needed for estimation.  

We address this through a novel formulation of spatio-temporal point processes.  
Specifically, we design the intensity as a deep mixture of experts, whose mixture weights are modeled by a deep neural network. 
This method, called \textsf{DMPP} (Deep Mixture Point Processes),  
enables us to incorporate unstructured contextual features (e.g., road networks and social/traffic event descriptions) into the predictive model,
and to automatically learn their complex effects on event occurrence. 
Moreover, this formulation yields a tractable optimization problem. 
Our mixture model-based approach permits the likelihood to be determined from tractable integration.  
Learning can be done with simple back-propagation. 

We conduct experiments on three real-world data sets from multiple urban domains and show 
that our \textsf{DMPP} consistently outperforms existing methods in event prediction tasks. 
The experiments also demonstrate that \textsf{DMPP} provides useful insights about why and under which circumstances events occur.  
By utilizing a recently developed self-attention mechanism \cite{Lu2017,Lin2017},  
\textsf{DMPP} helps us better understand how the contextual features influence event occurrence.  
Such insights could further aid policy makers in creating more effective strategies. 
The main contributions of this paper are as follows:
\begin{itemize}
\item We propose \textsf{DMPP}, a novel method for spatio-temporal event prediction. It accurately and effectively predicts spatio-temporal events by leveraging the contextual features, such as map images and social/traffic event descriptions, that impact event occurrence.

\item We integrate the deep learning approach into the point process framework.  
Specifically, we extract the intensity by using a deep mixture of experts, whose mixture weights are modeled by a deep neural network. 
This formulation allows us to utilize the information present in unstructured contextual features, and to automatically discover their complex effects on event occurrence, while at the same time yielding tractable optimization.

\item We develop an efficient estimation procedure for training and evaluating \textsf{DMPP}. 

\item We conduct extensive experiments on real-world data sets from three urban domains. With regard to event occurrence, the proposed method achieves better predictive performance than all existing methods on all data sets (Section 6).
\end{itemize}

\section{Related Work}
Point process is a general mathematical framework for modeling a sequence of events; it directly estimates the rate of event occurrence, 
by using explicit information about location and/or time.  
Early work mainly focused on the temporal aspect of events. 
The temporal Hawkes processes \cite{hawkes1971spectra} are a class of temporal point process models that can capture burst phenomena;  
in these models, the probability of future events is assumed to be strengthened by past events, with  the influence decaying exponentially over time.  
They have been used for analysing disease transmissions \cite{choi2015constructing}, 
financial transactions \cite{bacry2015hawkes,ait2015modeling}, 
terrorist attacks \cite{porter2012self}, 
social activities \cite{iwata2013discovering,farajtabar2014shaping}, 
search behaviors \cite{li2014identifying}, and so on. 
Recent studies have expanded its application to human mobility modeling.  
Wang {\sl et al.} (\citeyear{Wang2017a}) proposed Hawkes process variant to identify trip purpose.  
Du {\sl et al.} (\citeyear{Du2016}) presented a recurrent marked temporal point process (RMTPP) and 
demonstrated its effectiveness in predicting the timing of taxi pick-ups.  
Log Gaussian Cox process (LGCP) has been used to effectively model temporal events,  
such as wildfires \cite{Serra2014} and infrastructure failures \cite{Ertekin2015}, 
in which the logarithm of the intensity is assumed to be drawn from a Gaussian process.  
The spatio-temporal point process is a more general framework, and considers both spatial and temporal domains. 
The spatio-temporal self-exciting point processes, an extension of temporal Hawkes processes, have been used for modeling 
seismicity \cite{ogata1998space}, 
contagious diseases \cite{schoenberg2017recursive}, and crime incidents \cite{Mohler2011}, among other applications
The spatio-temporal LGCP has been applied to model wildfires \cite{Serra2014} and infrastructure failures \cite{Ertekin2015}.  

All these methods, however, have one fundamental limitation: 
they ignore contextual features even though they are known to influence event occurrence.
Human activities are largely influenced by environmental features, 
i.e., weather, geographical characteristics and traffic conditions. 
These features must be considered to accurately predict future events.
Their influence has been modeled by a special point process model, called the proportional hazards model \cite{Cox1992};  it treats 
the intensity rate as a function of covariates. 
One major limitation of this model is that it assumes that the contextual features create only linear effects.  
Most features have highly non-linear effects on real world event occurrence. 
The simplest solution is to fit non-linear functions, such as polynomials \cite{Hayano2011} and splines \cite{Giorgi2003}, to covariates. 
Unfortunately, their assumptions may be too restrictive to capture complex and intricate effects of contextual features.   
Also, this approach forces us to carefully choose or design the functional form of the covariates so that they accurately capture reality. 
However, in practice, how the contextual features influence event occurrence is largely unknown. 

This paper constructs a novel point process method called \textsf{DMPP}; it extends the spatio-temporal point process with a deep learning model.  
The pioneering work by \cite{Du2016,Xiao2017,Xiao2017a} is most related to our approach.  
However, they focus only on the temporal dynamics of event occurrence, 
and so ignore spatial dynamics.  
Also, those methods are optimized to predict the timing of the next event.  
Instead, we are interested in predicting longer event sequences.  
Moreover, none of these methods accept contextual features.

\section{Preliminaries}
In this section, before introducing our method, we first provide the necessary theoretical background to the point process. 

Point process is a random sequence of event occurrences over a domain. 
We assume here a sequence of events with known times and locations.  
Let ${\bf x} = (t,s)$ be the event written as the pair of time $t\in\mathbb{T}$ and location $s\in\mathbb{S}$,  
where $\mathbb{T}\times \mathbb{S}$ is a subset of $\mathbb{R}\times\mathbb{R}^2$. 
In the following, we denote the number of events falling in subset $A$ of $\mathbb{T}\times \mathbb{S}$ as $N(A)$.  
The general approach to identifying a point process is to estimate the ``intensity'' 
$\lambda({\bf x})$. The intensity $\lambda({\bf x})$ represents the rate of event occurrence in a small region, and is defined as 
\begin{align}
\lambda({\bf x}) = \lambda(t,s) \equiv \lim_{|dt|\rightarrow 0,|ds|\rightarrow 0} \frac{\mathbb{E}[N(dt \times ds)]}{|dt||ds|},
\end{align}
where $dt$ is a small interval around time $t$, $|dt|$ is its duration, $ds$ is a small region containing location $s$, and $|ds|$ is its area. 
$\mathbb{E}$ indicates an expectation measure.    
The functional form of intensity is designed to appropriately capture the underlying dynamics of event occurrence. 

Given a sequence of events $\mathcal{X}=\{{\bf x}_i = (t_i,{\bf s}_i)\}_{i=1}^N$, $t_i\in\mathbb{T}$ and ${\bf s}_i\in\mathbb{S}$,  
the likelihood is given by  
\begin{align}
p(\mathcal{X}|\lambda({\bf x})) = \prod_{i=1}^N \lambda({\bf x}_i) \cdot 
\exp{\bigg(-\int_{\mathbb{T}\times\mathbb{S}} \lambda({\bf x}) d{\bf x}\bigg)}. 
\end{align}

\section{Deep Mixture Point Processes}
This section presents the proposed method referred to as \textsf{DMPP} (Deep Mixture Point Processes). 
We first introduce the notations and definitions used in this paper.  
We then provide the model formulation of \textsf{DMPP} followed by parameter learning and prediction. 
The neural network architecture used in \textsf{DMPP} is detailed in the Appendix.  

\subsection{Problem Definition}
We introduce the notations used in this paper and formally define the problem of event prediction.  

Let $\mathcal{X}=\{{\bf x}_i = (t_i,{\bf s}_i)\}_{i=1}^N$ denote a sequence of events over space and time,  
where ($t_i$,${\bf s}_i$)$\in\mathbb{T}\times \mathbb{S} \in \mathbb{R}\times\mathbb{R}^2$ and $N$ is the total number of events known.  

Further, we are also given contextual information associated with the spatio-temporal region $\mathbb{T}\times \mathbb{S}$. 
Let $\mathcal{D}=A_1, A_2, ..., A_K$ be a set of contextual features, 
where $A_k$ is the $k$-th feature, and $K$ is the number of contextual features.  
Examples of the contextual features include weather, social/traffic event information and geographical characteristics.  
The social/traffic event information may be a collection of social/traffic event descriptions that include locations and times. 
In this case, $A_{\text{\tiny{\textbullet}}}$ is represented by a set of four-element tuples, each of which has the following format: \texttt{<time, latitude, longitude, description>}.  
Information about the geographical characteristics can be obtained from map images.  

Given the contextual features $\mathcal{D}$ up to time $T+\Delta T$, and the event sequence $\mathcal{X}$ up to time $T$, 
we aim to learn a predictor that:   
\begin{itemize}
\item predicts times and locations of events in the future time window $[T, T+\Delta T]$; 
\item predicts the number of events within any given spatial region and the time period in $[T, T+\Delta T]$, 
\end{itemize}
by leveraging $\mathcal{D}$ and $\mathcal{X}$.

\subsection{Model Formulation}
In this work, we construct a novel point process method for spatio-temporal event prediction that can incorporate unstructured contextual features 
such as map images and social/traffic event descriptions. 
Our point process intensity must be designed so that it is flexible enough to capture the highly complex effects of contextual features, 
while at the same time being tractable. 
Deep learning models have proven to be an extremely useful, especially in automatically extracting the meaningful information 
contained in the unstructured data including images and text descriptions. 
Inspired by this, we propose a novel formulation of point process model by integrating it with deep learning approach.  
The proposed method is referred to as \textsf{DMPP} (Deep Mixture Point Processes). 
In particular, we model the intensity by a neural network function that accepts contextual features as its input. 

 
{\bf Intensity function.}
We develop a flexible and computationally effective way of using kernel convolution to specify the intensity function.  
Formally, we design the intensity as a function of contextual features: \begin{align}\label{eq:conv}
\lambda({\bf x}|\mathcal{D}) = \int f\big({\bf u},{\bf Z}({\bf u};\mathcal{D});\theta\big) k({\bf x}, {\bf u}) d{\bf u},
\end{align}
where ${\bf u}=(\tau, {\bf r})$ for $\tau\in\mathbb{T}$ and ${\bf r}\in\mathbb{S}$, $k(\cdot, {\bf u})$ is a kernel function centered at ${\bf u}$. 
$f(\cdot)$ is any deep learning model that returns a nonnegative scalar, and $\theta$ denotes a set of the parameters of the deep neural network.  
${\bf Z}({\bf u},\mathcal{D})=\{Z_1({\bf u};A_1),...,Z_K({\bf u};A_K)\}$ is a set of the feature values at the spatio-temporal point ${\bf u}$, 
where $Z_k$ is defined as the operator to extract values of $k$-th feature at ${\bf u}$.  
As one example, social/traffic event descriptions can be represented by tuples of time, location and event descriptions. In this case, operator $Z$ outputs a list of social/traffic event descriptions 
scheduled within $[\tau-\Delta\tau, \tau+\Delta\tau]$ and located within a predefined distance, $||{\bf r} - {\bf r}' || < \Delta{\bf r}$, given ${\bf u}=(\tau,{\bf r})$.  
As another example, given a map image representing geographical characteristics, 
$Z$ returns the feature vectors (e.g., RGB values) of the map image around ${\bf r}$. 
The formulation of Equation (\ref{eq:conv}) is built upon a process convolution approach \cite{higdon2002space,lemos2009spatio,lee2008inference}; but we extend it so that the point process intensity  
accepts unstructured contextual features, by integrating it with a deep neural network.  
This extension enables us to integrate unstructured contextual data,  
and automatically learn their complex effects on event occurrence.  
Although being flexible and expressive, this intensity is intractable 
as it involves the integral of the neural network function $f(\cdot)$.   
Thus, by introducing $J$ representative points $\mathcal{U} = \{{\bf u}_j\}_{j=1}^J$ in the spatio-temporal region, 
we obtain a discrete approximation to Equation (\ref{eq:conv}): 
\begin{align}\label{eq:intensity}
\lambda({\bf x}|\mathcal{D}) = \sum_{j=1}^J f\big({\bf u}_j,{\bf z}_j;\theta\big) k({\bf x}, {\bf u}_j),
\end{align}
where each point ${\bf u}_j=(\tau_j, {\bf r}_j)$ consists of its time $\tau_j\in\mathbb{T}$ and location ${\bf r}_j\in\mathbb{S}$.  
Here we define ${\bf Z}({\bf u}_j; \mathcal{D})$ as ${\bf z}_j$,  
which represents the contextual feature vector associated with the $j$-th point ${\bf u}_j$. 
Consequently, the intensity is described as a mixture of kernel experts, in which mixture weights are modeled 
by a deep neural network whose inputs are contextual features. 
The resulting model yields the automatic learning of their influences as well as making the learning problem tractable (discussed in Section 4.3).

{\bf Configuration of representative points. }
The set of representative points is structured as follows.  
We first introduce $M$ discrete points placed uniformly along time axis within $[0, T+\Delta T]$ to define time points $\mathcal{T}$: 
$0 = \tau'_1 < ... < \tau'_{M} = T+\Delta T$. 
Similarly, we set $L$ discrete points within the spatial region to define space points $\mathcal{S}$:  
${\bf r}'_1, ..., {\bf r}'_{L}$, where ${\bf r}'_l \in\mathbb{S}$. 
The set of representative points is defined by the Cartesian product of the space and time points: 
$\mathcal{U} = \{(\tau, {\bf r}) \mid \tau\in\mathcal{T} \wedge {\bf r}\in\mathcal{S}\}$. Therefore $J=ML$. 
There are some options in locating the representative points, either fixing them or optimizing them in terms of spatial coordinates.  
In this paper, we choose the former and fix them on a regular grid, as simplifies the computation. 
Note that the number of representative points, $J$, determines the trade-off between approximation accuracy and computation complexity. 
Larger $J$ improves approximation, while reducing computational cost.  
A sensitivity analysis of the impact of $J$ is given in the experimental section. 

{\bf Kernel function. }
We can make various assumptions as to the kernel function $k({\bf x}, {\bf u}_j)$. 
For example, we can use a Gaussian kernel: \begin{align}
k({\bf x},{\bf u}_j) = \exp{\big(- ({\bf x}-{\bf u}_j)^\top \Sigma^{-1} ({\bf x}-{\bf u}_j)\big)}, \end{align}
where $\Sigma$ is \textcolor{black}{a $3\times 3$ covariance matrix (bandwidth) of the kernel.} 
Other kernel functions, such as Matern, sigmoid, periodic (trigonometric), and compactly supported kernels \cite{wendland1995piecewise} are viable alternatives.  

{\bf Neural network model.}
The neural network model $f(\cdot)$ can be designed to suit the input data. 
We consider the general case wherein image features (e.g., map images) and text features (e.g., social/traffic event descriptions) are available. 
We propose an attention network architecture that fully exploits visual and textual information. 
The proposed architecture consists of three components: 
{\sl an image attention network} to extract image features, {\sl a text attention network} to encode text features, and {\sl a multimodal fusion module}. We construct {\sl the image attention network} by combining a CNN with the spatial attention model proposed by \cite{Lu2017}.  
We design {\sl the text attention network} on a CNN designed for sentences~\cite{kim2014convolutional} and an attention mechanism \cite{Lin2017}.  
The structured texts are split into words and then processed by the text attention network.  
Lastly, the extracted features from images and texts are fused into a single representation via {\sl the multimodal fusion module}, 
and input to the intensity function. 
We detail each component of the neural network in Appendix A.   

\subsection{Parameter Learning}
Given a list of observed events up to time $T$ (total of $N$ events) $\mathcal{X}$, 
the logarithm of the likelihood function is written as 
\begin{align}\label{eq:like}
\log{p(\mathcal{X}|\lambda({\bf x}))} &= \sum_{i=1}^N \log\lambda({\bf x}_i|\mathcal{D}) - \int_{\mathbb{T}\times\mathbb{S}} \lambda({\bf x}|\mathcal{D}) d{\bf x} \nonumber \\&= \sum_{i=1}^N \log \sum_{j=1}^J f({\bf u}_j,{\bf z}_j;\theta) k({\bf x}_i,{\bf u}_j) \nonumber \\ & - \sum_{j=1}^J f({\bf u}_j,{\bf z}_j;\theta) \int_{\mathbb{T}\times\mathbb{S}} k({\bf x},{\bf u}_j) d{\bf x}, 
\end{align}
where $\mathbb{T}\times\mathbb{S}$ is the domain of the observation. 
Notably, the above log-likelihood can be solved tractably with integratable kernel functions.   
Our mixture model-based approach with representative points allows the neural network model $f(\cdot)$, which cannot be integrated analytically in general, 
to be moved outside the integral.  
This permits us to use the simple back-propagation algorithm. 
For many well-known kernel functions, such as Gaussian, polynomial, the integral of the second term is written as closed-form solutions or approximations.  
In the case of the Gaussian kernel, it is described by an error function.  
During the training phase, we adopt mini-batch optimization.  
Over the set of indices selected in a mini-batch $\mathcal{I}$, by normalizing the first term in Equation (\ref{eq:like}), 
the objective function can be written as 
\begin{align}\label{eq:like1}
\log{p(\mathcal{X}|\lambda({\bf x}))} 
&= \frac{N}{|\mathcal{I}|} \sum_{i\in\mathcal{I}} \log \sum_{j=1}^J f({\bf u}_j,{\bf z}_j;\theta) k({\bf x}_i,{\bf u}_j) \nonumber \\
& - \sum_{j=1}^J f({\bf u}_j,{\bf z}_j;\theta) \int_{\mathbb{T}\times\mathbb{S}} k({\bf x},{\bf u}_j) d{\bf x}, 
\end{align}
where $|\mathcal{I}|$ denotes the mini-batch size.  
We apply back-propagation to find all the model parameters, $\Theta = \{\Sigma,\theta\}$, that maximize the above log-likelihood,     
by taking the derivative of Equation (\ref{eq:like1}) 
w.r.t. kernel parameter $\Sigma$ and neural network parameters $\theta$.  
 
\subsection{Prediction}
Here we present a procedure for future event prediction.  

We denote representative points within the test period $\mathbb{T}^{\ast} = (T,T+\Delta T]$ as $\mathcal{U}^{\ast} = \{(\tau,{\bf r}) \mid T < \tau \leq T + \Delta T\}\subset\mathcal{U}$.  
Given the learned parameters of the neural network, $\hat{\theta}$, 
we first calculate $f({\bf u}_j,{\bf z}_j; \hat{\theta})$ for each representative point.  
Using the set of estimated functions $\{f({\bf u}_{j},{\bf z}_{j};\hat{\theta})\}_{{\bf u}_j\in\mathcal{U}^{\ast}}$ 
and the estimated kernel parameter $\hat{\Sigma}$, 
we derive intensity $\hat{\lambda}({\bf x})$ for the test period based on Equation (\ref{eq:intensity}). 

Given the sequence of events observed in the test period $\mathbb{T}^{\ast}$, $\mathcal{D}=\{{\bf x}_{N+1}, ..., {\bf x}_{N+n}\}$,  
analogous to Equation (\ref{eq:like}), the log-likelihood for the test data is calculated as  
\begin{align}
\mathcal{L}^{\ast} =& \log{p(\mathcal{D}|\hat{\lambda}({\bf x}))} 
= \sum_{i=N+1}^{N+n} \log \sum_{{\bf u}_j\in\mathcal{U}^{\ast}} f\big({\bf u}_j,{\bf z}_j;\hat{\theta}\big) k({\bf x}_i,{\bf u}_j) \nonumber \\
& - \sum_{{\bf u}_j \in \mathcal{U}^{\ast}} f\big({\bf u}_j,{\bf z}_j;\hat{\theta}\big) \int_{\mathbb{T}^{\ast}\times\mathbb{S}} k({\bf x},{\bf u}_j) d{\bf x}.  
\end{align}

The point process model can be used to predict the expected number of events. 
The number of events is derived by integrating the estimated intensity over specific time period $P\subset\mathbb{T}^{\ast}$ 
and region of interest $Q\subset\mathbb{S}$ such that 
\begin{align}\label{eq:int}
N(P\times Q) = \int_{P\times Q} \hat{\lambda}({\bf x}) d{\bf x} = \sum_{{\bf u}_j \in \mathcal{U}^{\ast}} f\big({\bf u}_j,{\bf z}_j;\hat{\theta}\big) \int_{P\times Q} k({\bf x},{\bf u}_j) d{\bf x}, 
\end{align}
where $N(A)$ is the number of events that fall into subset $A$. 
As discussed in Section 4.3, the above integral has a tractable solution.

\section{Experiments}
In this section, we use real-world data sets from different domains to evaluate the predictive performance of our model. 

\subsection{Data Sets}
\subsubsection{Event data}
We used three event data sets from different domains collected in New York City and Chicago  
from Jan 1, 2016 to April 1, 2016 (the observation period is 13 weeks). 
The details are as follows. 

{\bf NYC Collision Data.} New York City vehicle collision (NYC Collision) data set 
contains $\sim$ 32 thousand motor vehicle collisions.  
Every collision is recorded in the form of time and location (latitude and longitude coordinates). 

{\bf Chicago Crime Data.} Chicago crime data set is a collection of reported incidents of crime that occurred in Chicago; it   
contains $\sim$ 13 thousand records, each of which shows time, and latitude and longitude of where the crime happened.  

{\bf NYC Taxi Data.} New York City taxi pick-up (NYC Taxi) data set consists of $\sim$ 30 million pick-up records in New York City 
collected by the NYC Taxi and Limousine Commission (TLC). 
Each record contains pick-up time, latitude and longitude coordinate. 
To reduce data size, we randomly selected 100 thousand events for our experiment. 

For the 13-week observation period, we selected the last seven days as the test set, 
the last seven days before the test period as the validation set,  
and used the remaining data as the training set.
Thus, $T=120960$min and $\Delta T=10080$min. 

\subsubsection{Urban contextual data}
We used the following urban data as the contextual features.  

{\bf Map Image.} As the image features, we used {\sl the map image} of the cities acquired from OpenStreetMap (OSM) database
\footnote{Map data copyrighted OpenStreetMap contributors and available from https://www.openstreetmap.org}. 
For each representative point ${\bf u}_j = (\tau_j,{\bf r}_j)$, we extracted the image around ${\bf r}_j$ (i.e., about 300m$\times$ 500m square grid space) and used its RGB vector as the input of the \textcolor{black}{{\sl image attention network}}. 

{\bf Social/Traffic Event Description.} 
We collected {\sl traffic events} (e.g., major street construction works and street events) 
and {\sl social events} (e.g., sports events, musical concerts and festivals) in New York City as held by the 511NY website\footnote{https://www.511ny.org} during Jan 2016 through April 2016.  
The 13 week period contained a total of 8,968 descriptions.  
Each social/traffic event record contains a description of the event, as well as its start time $t_s$, end time $t_e$ and location (latitude and longitude coordinates). 
For each representative point ${\bf u}_j=(\tau_j,{\bf r}_j)$, we extracted social/traffic event descriptions satisfying $t_s<\tau_j<t_e$ and located within a predefined distance, $\Delta r$ from ${\bf r}_j$. 
If the point contains more than one sentence, we selected the spatially closest one.  
If the point contains no sentences, we used dummy variables. 
We used their descriptions, a sequence of 1-of-K coded word vectors, as the input of the text network. 
In this paper, we set $\Delta r$ to the walking distance of 620m, following \cite{Chen2016}.

\subsection{Experimental Setup}
Hyper-parameters of each model are tuned by grid-search on the validation set. 
For \textsf{DMPP}, we used the Adam algorithm \cite{Kingma2014} as the optimizer,  
with $\beta_1 = 0.01$, $\beta_2 = 0.9$, and learning rate of 0.01. 
For {\sl the multimodal fusion module} of \textsf{DMPP}, we tuned the hyper-parameters as follows:  
layer size $n_l$ in \{1,2,3,4\}; number of units per layer $n_u$ in \{16, 32, 64\}. 
The mini-batch size $|\mathcal{I}|$ is selected from the set \{8, 16, 32\}. 
Following the prior settings in \cite{Du2016}, we also applied L2 regularization with $\lambda$=0.001 in both models.  
The number of representative points are tuned on the validation set in terms of the number of time points, $M$, and the number of space points, $L$. 
The tested combinations were $M=\{24, 28, 168\}$ and $L=\{4,8,10,12\}$.  
We used three kinds of kernel functions: uniform, Gaussian, compactly supported Gaussian (the definitions are provided in Appendix B).  
Also, we explored various map styles: OSM default (the original map of Figure \ref{fig:attention_map}a), Watercolor (the original map of Figure \ref{fig:attention_map}b), Greyscale. 
The best settings for \textsf{DMPP} are given in the corresponding section.  
The default settings for each component of the neural network are described in Appendix A.

\begin{figure*}[t]
\centering
\includegraphics[height=1.65in]{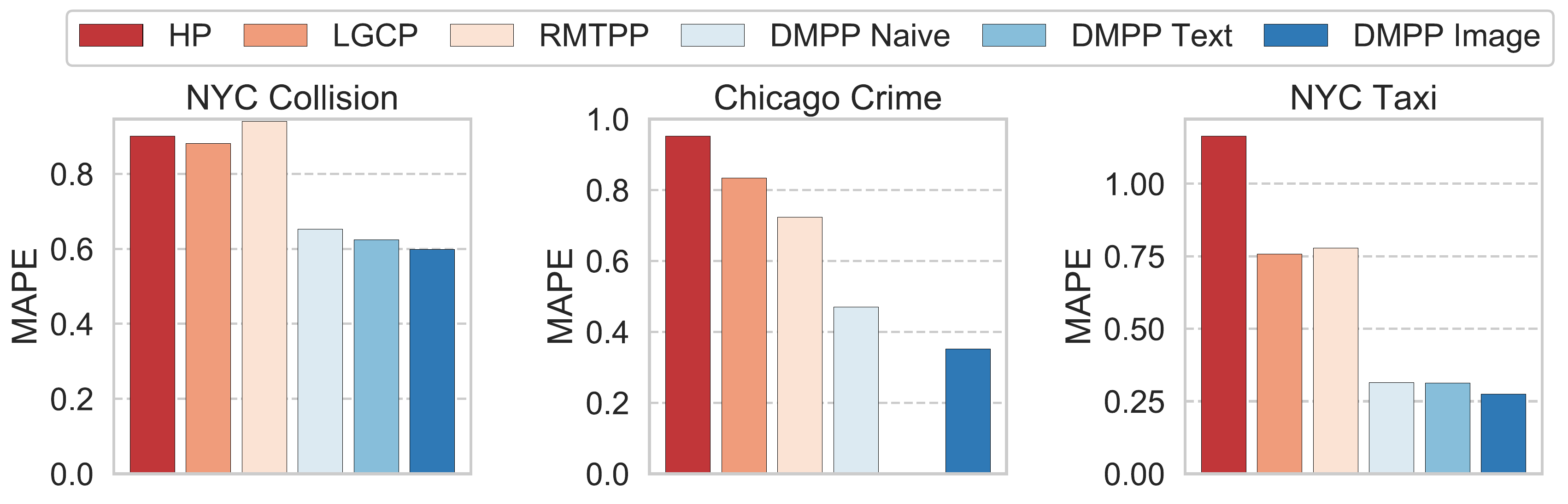}
\caption{
MAPE for event number prediction from six methods on three data sets:  
NYC Collision data (left); Chicago Crime data (middle); NYC Taxi data (right). Lower is better. 
DMPP is the proposed method.  
The error bars are omitted as the deviations are negligible.  
}\label{fig:mae}
\end{figure*}

\subsection{Evaluation Metrics}
We evaluated the predictive performance using two metrics: 
{\bf LogLike} (predictive log-likelihood) and {\bf MAPE} (Mean Absolute Percentage Error).  
For the first metric, given the learned model,  
we calculated log-likelihood on the test data ({\bf LogLike}) for each event as $\mathcal{L}^{\ast}/N_t$, where $\mathcal{L}^{\ast}$ is the test log-likelihood 
defined by Equation (7) and $N_{\text{t}}$ is the number of test events. 
{\bf MAPE} is used to evaluate the performance of event number prediction; it is defined as the absolute difference between the predicted number of events and the actual number: 
$\text{MAPE} = \sum_{r=1}^{N_r} \sum_{t=1}^{N_b} |n_{r,t} - \hat{n}_{r,t}| / n_{r,t}$, 
where $n_{r,t}$ is the number of events observed in the $r$-th grid cell and $t$-th time interval,   
and $\hat{n}_{r,t}$ is the corresponding prediction. $N_r$ is the number of grid cells and  
$N_b$ is the number of time bins for which predictions are made.  
In our experiment, we partitioned the region of interest using a $10\times 10$ uniform grid,   
and divided the test period (seven days) into 14 time bins with a fixed uniform interval of 12 hours. 
Therefore $N_r=100$ and $N_b=14$.  
For \textsf{DMPP}, we predicted the number of events for each pair of spatial grid cell and future time bin, using Equation (6). 

\subsection{Comparison Methods}
We compared the proposed model and its variants with three existing methods. 

\begin{itemize}\item \textsf{HP} (Homogeneous Poisson process): 
The intensity is assumed to be constant over space and time: $\lambda({\bf x})=\lambda_0$.  
The optimization can be solved in closed form. 
\item \textsf{LGCP} (Log Gaussian Cox process) \cite{diggle2013spatial}: \textsf{LGCP} is a kind of Poisson process with varying intensity, 
where the log-intensity is assumed to be drawn from a Gaussian process (See Appendix C). 
For \textsf{LGCP}, we performed the comparison only on event number prediction, since the log-likelihood of this model is computationally intractable. 
The inference is based on the Markov chain Monte Carlo (MCMC) approach (see \cite{Panik2010} for details). 
For event number prediction, we sampled events from \textsf{LGCP} using the thinning method \cite{lewis1979simulation}, 
and compared the aggregated number of events within predefined spatial regions and time periods with the ground truth. 
\item \textsf{RMTPP} (Recurrent Marked Temporal Point Process) \cite{Du2016}:  
\textsf{RMTPP} uses RNN to describe the intensity of the marked temporal point process; it assumes a partially parametric form for the intensity, and can capture temporal burst phenomena. This model is primarily intended to model event timing; to allow comparison, we mapped latitude and longitude values into location names and treating them as marks, using Neighborhood Names GIS data
\footnote{NYC Neighborhood Names GIS data, https://data.cityofnewyork.us/City-Government/Neighborhood-Names-GIS/99bc-9p23}$^{,}$\footnote{Chicago Neighborhood Names GIS data, https://data.cityofchicago.org/Facilities-Geographic-Boundaries/Boundaries-Neighborhoods/bbvz-uum9} (details are provided in Appendix C).  
The following hyper-parameters are tuned on the validation set: Number of units per layer in \{16, 32, 64\} and mini-batch size in \{8, 16, 32\}. 
The optimal settings are given in Appendix C.  
Note that the likelihood of \textsf{RMTPP} is for the location names, not for latitude and longitude values. 
\textcolor{black}{For event number prediction, we sequentially predicted next event (See Appendix C). 
The predicted location names are mapped to latitude and longitude coordinates by simply using their centroids; and then  
the generated events are aggregated into counts.} 
\end{itemize}

We introduce three variants of \textsf{DMPP} below.  
\begin{itemize}\item \textsf{DMPP} {\sl Naive}: The simplest variant of \textsf{DMPP}, it does not incorporate any contextual features.  
\textcolor{black}{The neural network of \textsf{DMPP} accepts the location and time of each representative point $\textbf{u}_j$.}
\item \textsf{DMPP} {\sl Image}/{\sl Text}: The \textsf{DMPP} variants that incorporate either map images or the social/traffic event descriptions,  
\textcolor{black}{as well as the locations and times of the representative points. }
\end{itemize}

\subsection{Quantitative results}
\begin{table}[t]
\caption{Comparison of the proposed method and its variants to the baselines. 
The number indicates the predictive log-likelihood per event (\textbf{LogLike}) for the test data.  
Higher is better. }
\begin{tabular}{cccc} \\ \toprule
      & NYC Collision & Chicago Crime & NYC Taxi \\ \midrule
\textsf{HP} & 8.232 & 9.384 & 10.473 \\ \midrule
\textsf{DMPP} {\sl Naive} & 8.296 & 9.614 & 11.311 \\ 
\textsf{DMPP} {\sl Text} &  8.297 & NA & 11.318 \\
\textsf{DMPP} {\sl Image} & {\bf 8.409} & {\bf 9.621} & {\bf 11.333} \\ \bottomrule
\end{tabular}\label{tab:like}
\end{table}
\begin{figure*}[t]
\centering
\includegraphics[height=4cm]{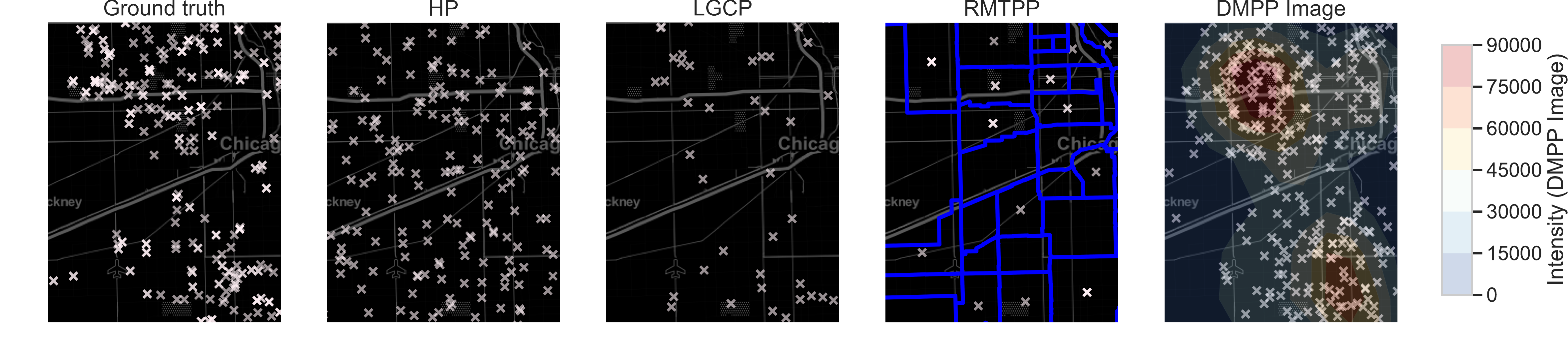}
\caption{ Events generated by four of the implemented methods for Chicago Crime data in Central and West Chicago between \textcolor{black}{0:00 am and 24:00 pm, on Mar 31th.} 
The cross markers (x) denote the events generated by simulations.  
In the fourth plot, the blue lines denote the location boundaries used for \textsf{RMTPP} in the experiment.  
In the right-most plot, we overlaid the estimated intensity for \textsf{DMPP} {\sl Image} at \textcolor{black}{Mar 31th 12:00 am. } 
}\label{fig:simulated}
\end{figure*}
Figure \ref{fig:mae} shows the overall MAPE of the six different methods on the three data sets for event number prediction. 
In this figure, the error bars are omitted as the deviations are negligible.  
The results indicate the superiority of our approach. 
\textsf{HP} performs worse than the other methods across almost all data sets,  
as it does not consider the spatio-temporal variation of the rate of event occurrence.  
We can see this in Figure \ref{fig:simulated}, which depicts events generated by four different methods from the Chicago Crime data. 
We simulated events with the thinning algorithm \cite{lewis1979simulation}, using the learned intensity of each method.   
\textsf{LGCP} presents better performance for NYC Collision and NYC Taxi data, as it captures spatio-temporal variations.  
\textsf{RMTPP} achieves relatively better performance than \textsf{LGCP} only for the Chicago Crime data.  
The result suggests that the assumption of \textsf{RMTPP}, the temporal burst phenomena, holds for Chicago Crime data, 
but not for NYC Collision data and NYC Taxi data.  
Even our simple model, \textsf{DMPP} {\sl Naive}, largely surpasses all existing methods. 
The result implies that the parametric assumptions of the existing methods are too restrictive, and do not capture real urban phenomena. 
Also, \textsf{RMTPP} intensity is influenced by all the past events, regardless of how spatially far away, 
so it is not suited for spatio-temporal events.  
The differences between \textsf{DMPP} {\sl Naive} and the best among the existing methods are significant (two-sided t-test: p-value$<0.01$) for all data sets.  
\textsf{DMPP} {\sl Naive} 
outperforms \textsf{LGCP} in terms of MAPE by 0.229 for the NYC Collision data,  
from 0.778 to 0.315 for the NYC Taxi data. \textsf{DMPP} {\sl Naive} outperforms \textsf{RMTPP} in terms of MAPE by 0.254 for the Chicago crime data.  
\textsf{DMPP} {\sl Text} further improves 
\textsf{DMPP} {\sl Naive} to 0.624 for the NYC Collision data, 
to 0.312 for NYC Taxi data.  
We can clearly see that \textsf{DMPP} {\sl Image} offers significantly improved prediction performance.  
From these results, we can conclude that considering urban contexts is very effective in improving event prediction performance. The results also suggest that our proposal, \textsf{DMPP}, effectively utilizes the information provided by urban contexts.  

Table \ref{tab:like} lists the LogLike values (predictive log-likelihood) of the four different methods for the three data sets, 
i.e., NYC Collision (vehicle collision) Data, Chicago Crime Data and NYC Taxi (taxi pick-up) Data. 
Note that \textsf{DMPP} {\sl Text} is not applicable to Chicago Crime data, as no text data is available for Chicago. 
The proposal, \textsf{DMPP}, outperforms \textsf{HP}. 
Even the simplest variant of \textsf{DMPP}, \textsf{DMPP} {\sl Naive} explains the observed event sequences better than these existing methods,  
which demonstrates the expressiveness of \textsf{DMPP}.   
\textsf{DMPP} {\sl Text} also outperforms \textsf{HP}. 
\textsf{DMPP} {\sl Image} achieves the best performance among all methods.  
This again shows the effectiveness of incorporating the urban contexts and our point process formulation with the deep neural network. 

\begin{figure*}[t]
\centering
\subfigure[Number of time points]{\centering\includegraphics[height=1.25in]{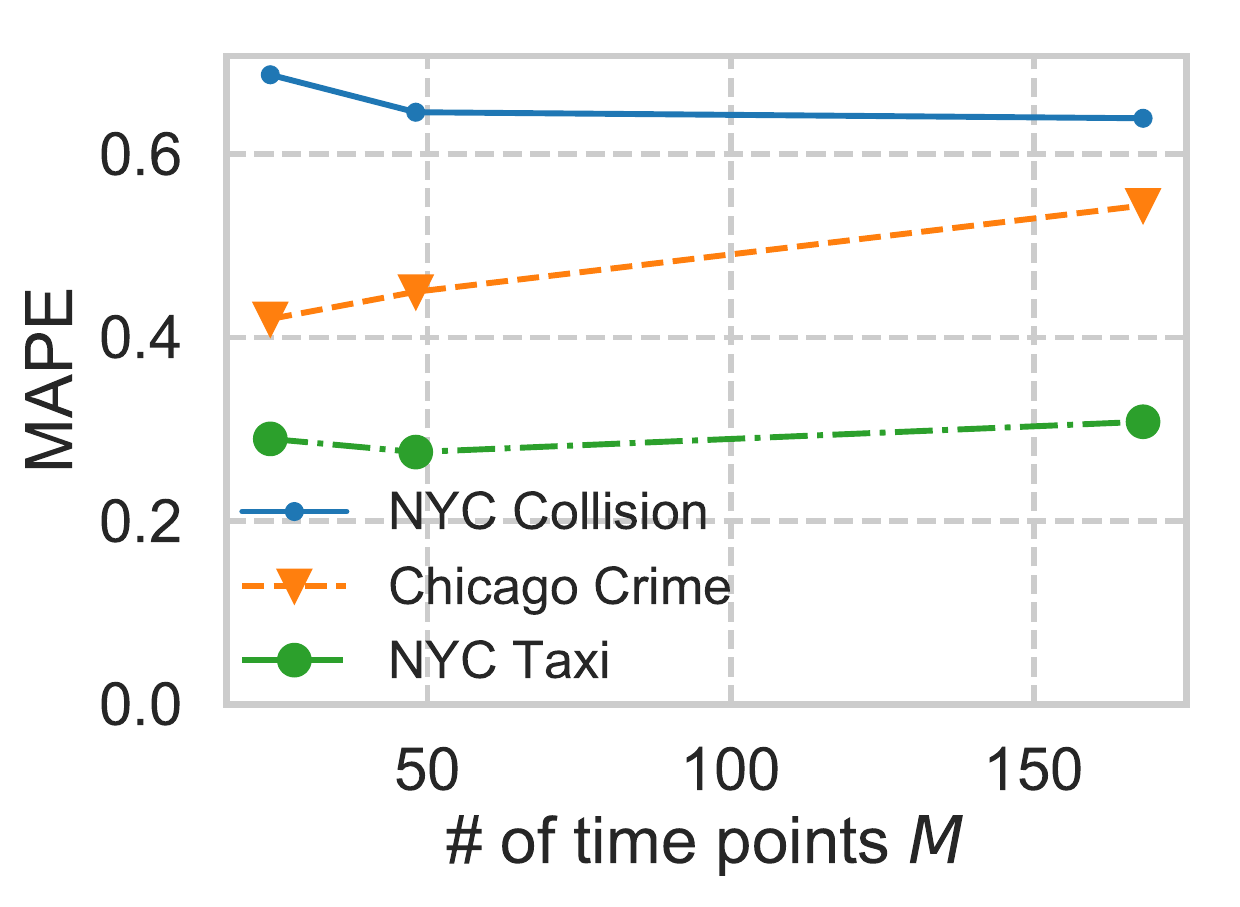}}\hspace{0.035cm}
\subfigure[Number of space points]{\centering\includegraphics[height=1.25in]{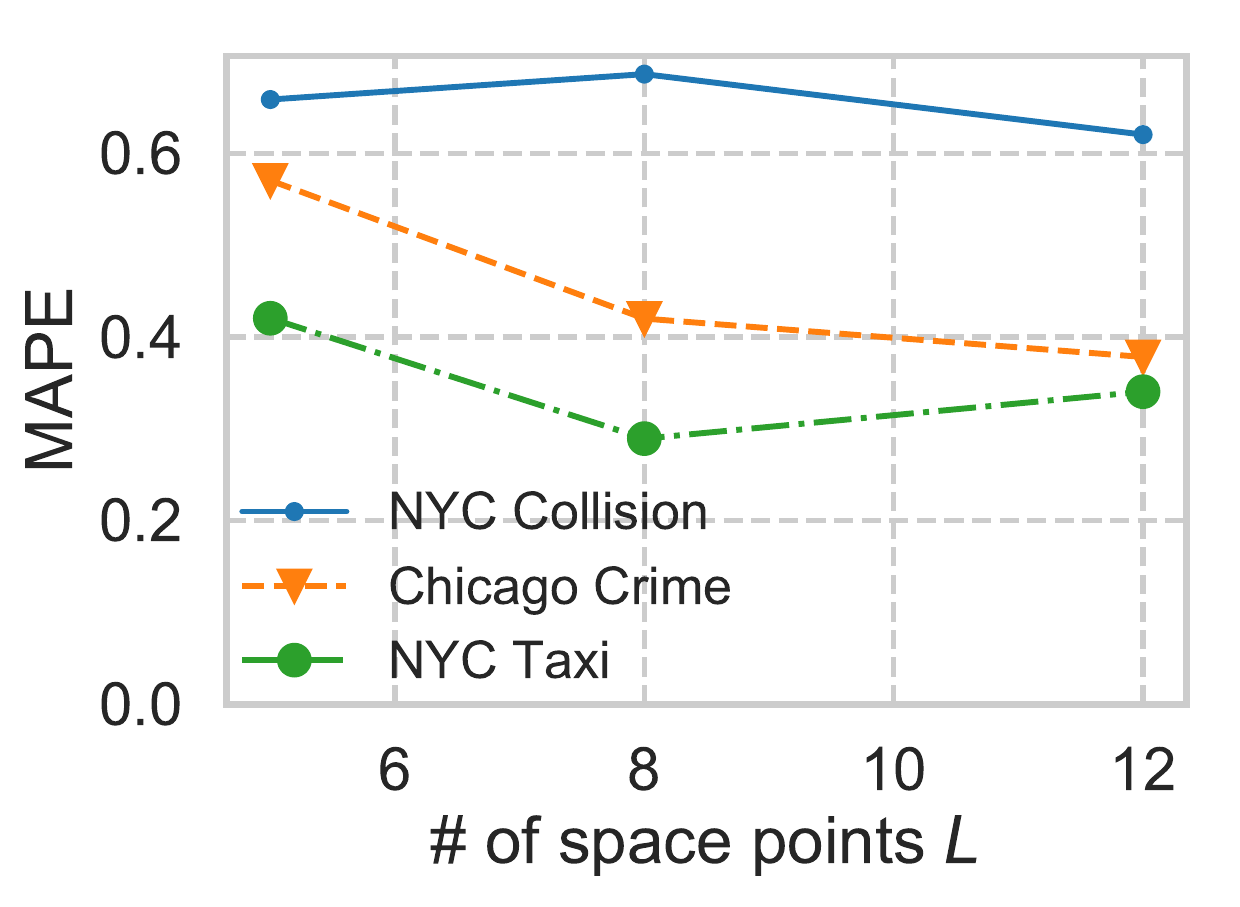}}\hspace{0.035cm}
\subfigure[Kernel function]{\centering\includegraphics[height=1.25in]{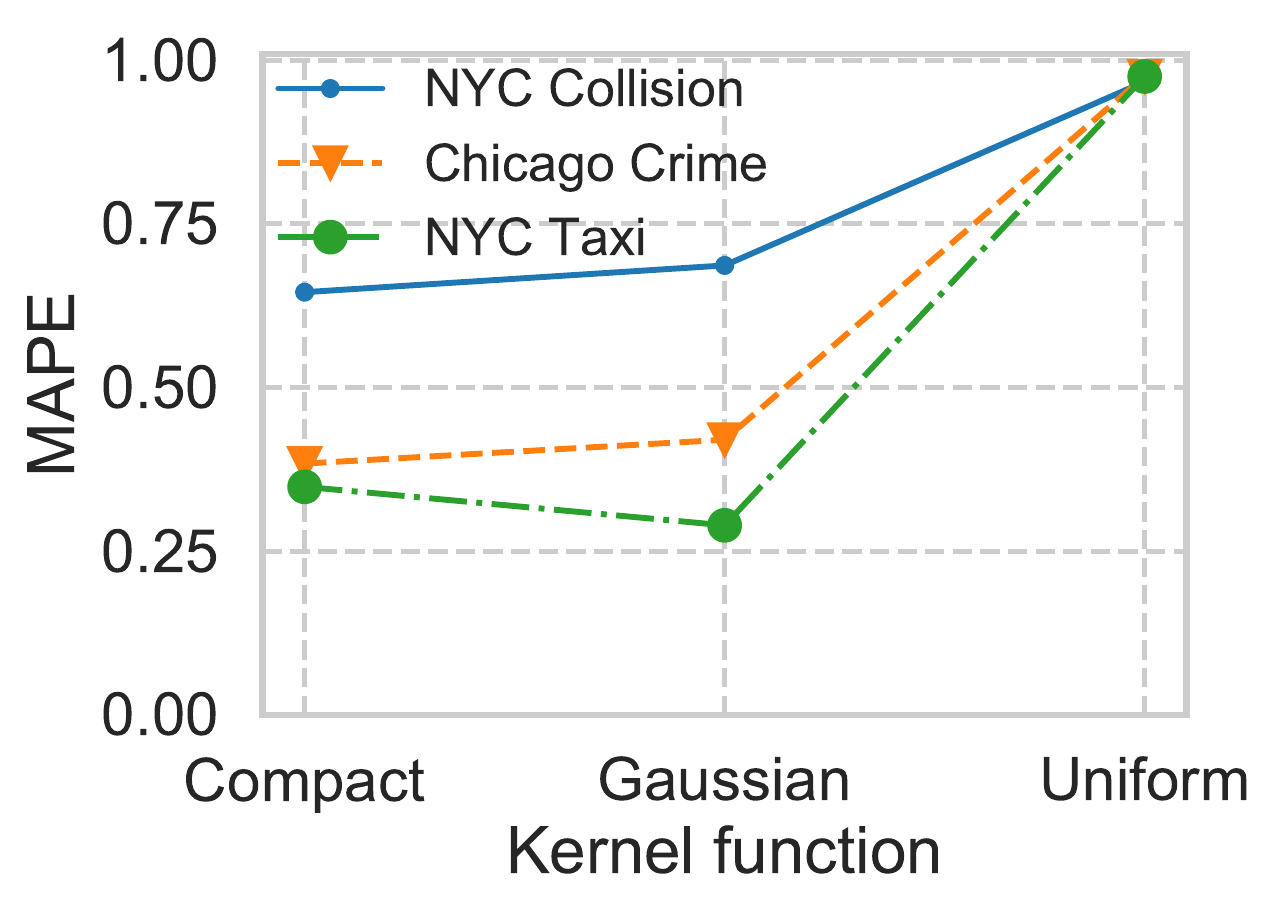}}\hspace{0.035cm}
\subfigure[Map style]{\centering\includegraphics[height=1.25in]{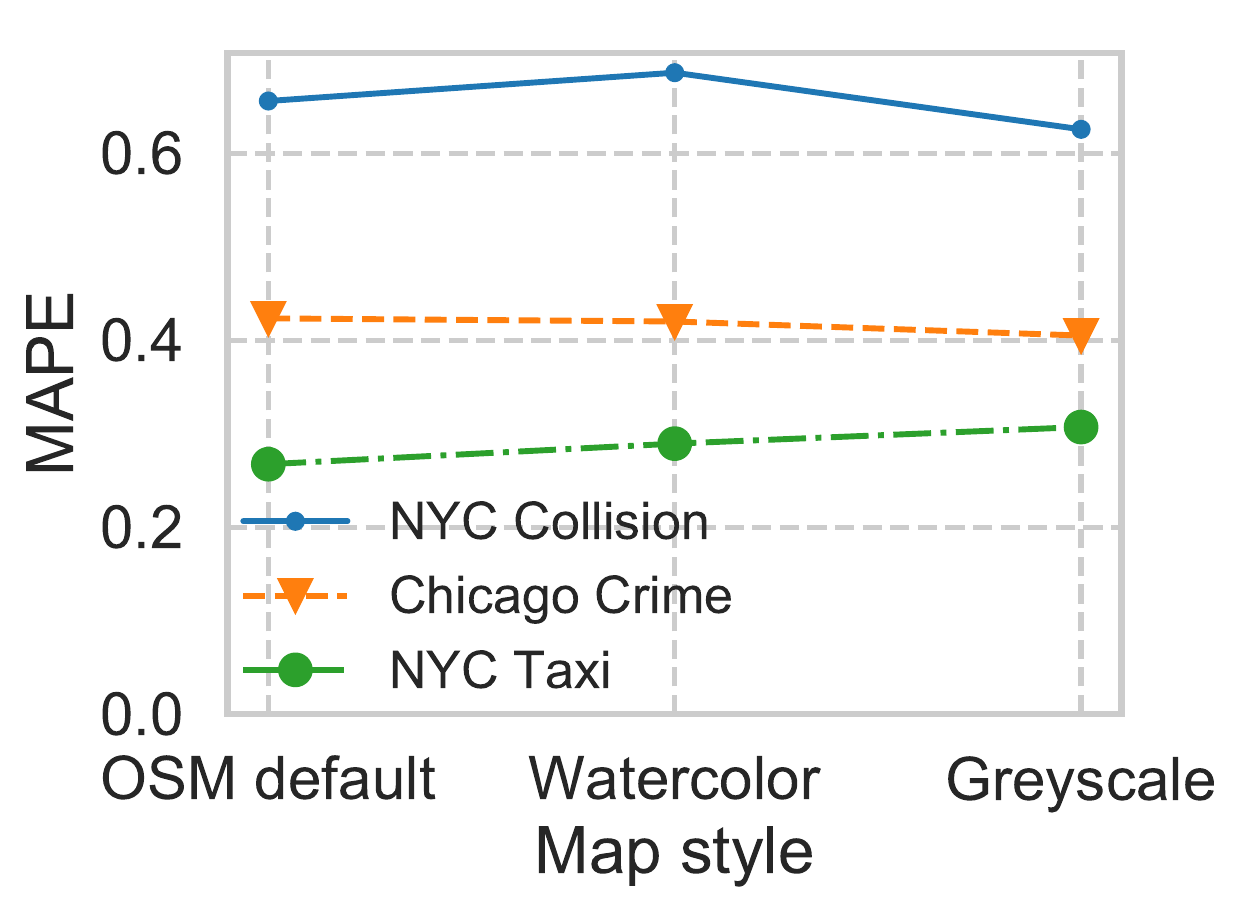}} \\
\subfigure[Layer size]{\centering\includegraphics[height=1.25in]{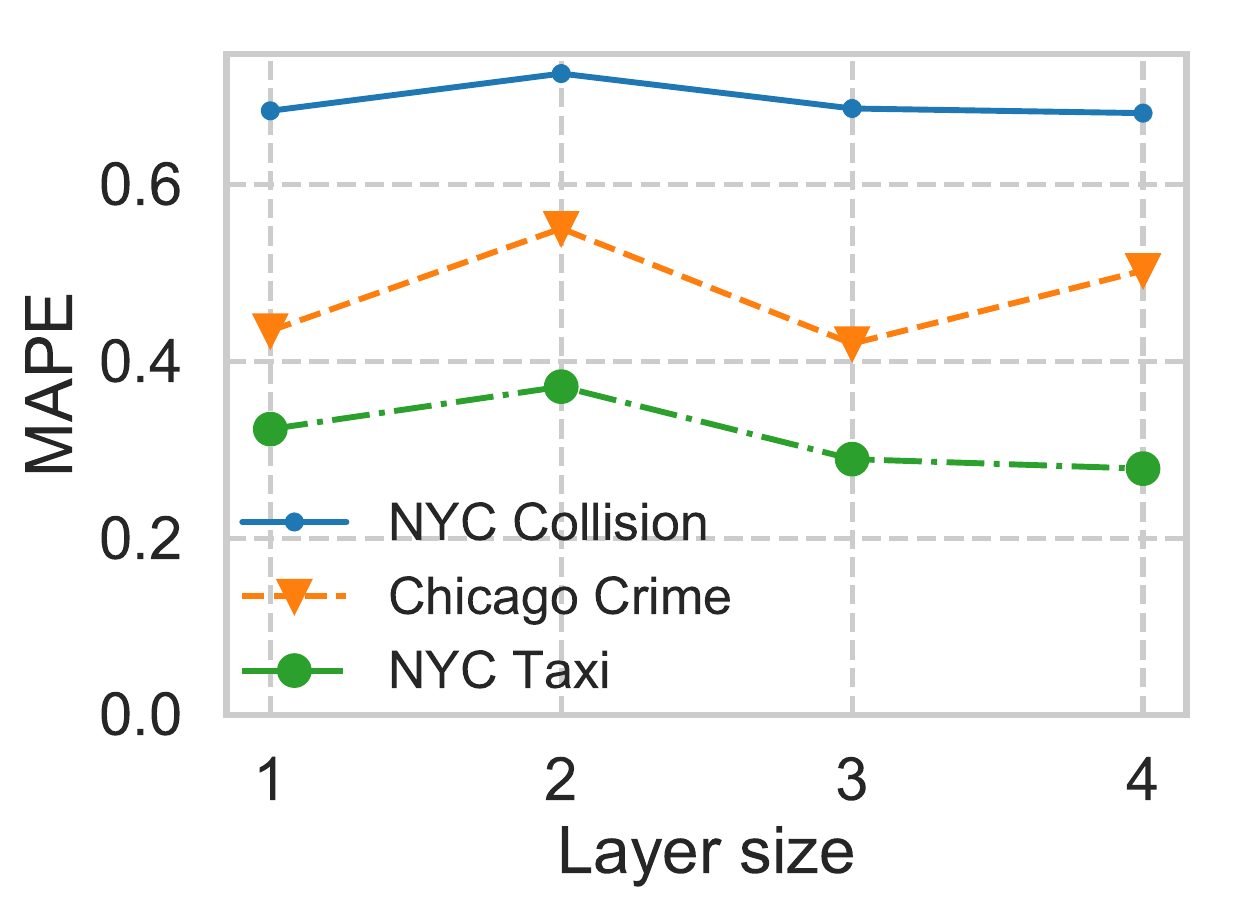}}\hspace{0.035cm}
\subfigure[Number of units per layer]{\centering\includegraphics[height=1.25in]{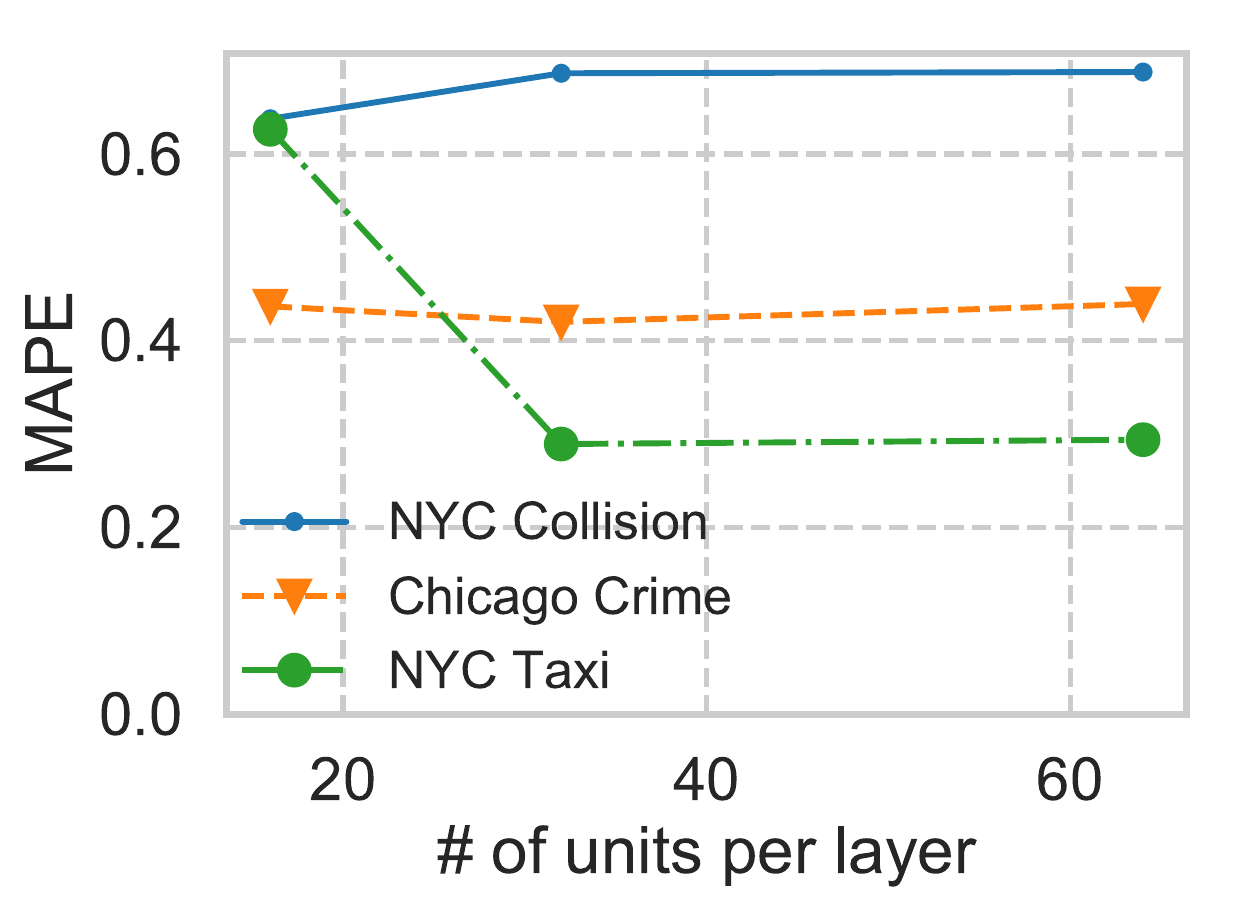}}\hspace{0.035cm}
\subfigure[Batch size]{\centering\includegraphics[height=1.25in]{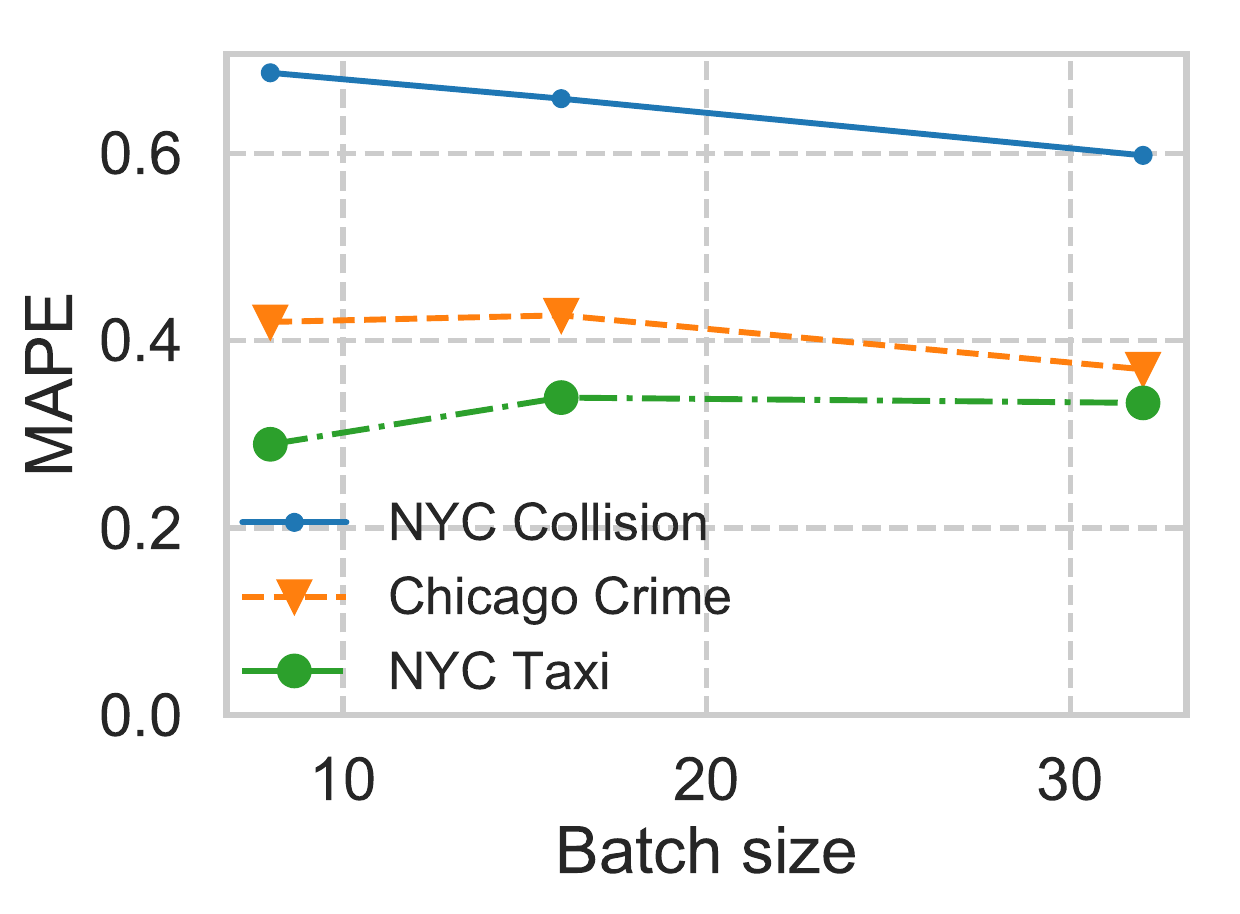}} 
\caption{
Impact of hyper-parameters on MAPE performance. 
}\label{fig:param}
\end{figure*}

\subsubsection{Sensitivity study}
Here we analyze the impact of parameters on \textsf{DMPP}, 
including (1) number of representative points; (2) kernel function (3) map style; and (4) neural network structure.  

\textbf{Number of Representative Points. } 
Figure \ref{fig:param}a and \ref{fig:param}b show the impact of the numbers of representative points on the performance of \textsf{DMPP} {\sl Image}. 
Figure \ref{fig:param}a shows that MAPE is slightly improved when 
\textcolor{black}{$M=168$ for NYC Collision data, $M=48$ for Chicago Crime data, 
and $M=24$ for NYC Taxi data.} 
As shown in Figure \ref{fig:param}b, the prediction performance of \textsf{DMPP} 
generally tends to increase with number of time points $M$.
 Overall, \textsf{DMPP} is moderately robust to variations in the numbers of representative points.  
In this experiment, we fixed the network depth $n_l$ to 4, the number of units per layer $n_u$ to 32 and batch size $|\mathcal{I}|$ to 16.  

\textbf{Choice of Kernel Function. }
Figure \ref{fig:param}c presents the prediction results with three kernel functions: Uniform, Gaussian, compactly supported Gaussian\textcolor{black}{ (definitions are provided in Appendix B)}.  
We can observe that the compactly supported kernel offers similar accuracy to the Gaussian kernel, while affording a computational advantage (See Appendix B). 
The uniform kernel performs worst.  
Throughout this paper, we use the compactly supported Gaussian kernel as the default setting. 
The hyper-parameters were set to $n_l=4$, $n_u=32$, $|\mathcal{I}|=16$, $M=24$ and $L=20$ in this experiment.  

\textbf{Choice of Map Style. }
Figure \ref{fig:param}d demonstrates the effect of map style.   
The predictive performance appears to be insensitive to the style of map images. 
For NYC Taxi data, MAPE is slightly improved when using the OSM default style. 
This may because the OSM default map distinguishes minor and major roads by color 
\textcolor{black}{(as shown in the original map image of Figure \ref{fig:attention_map}b).} 
As the default setting, we use OSM default style for NYC Collision and NYC Taxi data, and Watercolor for Chicago Crime data.  
In this experiment, we set $n_l=4$, $n_u=32$, $|\mathcal{I}|=16$, $M=24$ and $L=20$.  

\textbf{Network Structure. } 
We show the impact of network structures in Figure \ref{fig:param}e-\ref{fig:param}g. 
\textcolor{black}{The prediction performance slightly improves when layer size is 4, for NYC Collision data and NYC Taxi data, layer size 3 for Chicago Crime data.  
\textsf{DMPP} performs robustly for large number of units $n_u>=32$ across all the data sets.  
The prediction accuracy is likely to become better for larger batch size (Figure \ref{fig:param}f). }
In this experiment, we fix $M=24$ and $L=12$, respectively.  
The optimal value of network hyper-parameters correspond to \textcolor{black}{
$n_l=4$, $n_u=16$, $|\mathcal{I}|=32$ for NYC Collision data, 
$n_l=3$, $n_u=64$, $|\mathcal{I}|=32$ for Chicago Crime data, 
$n_l=4$, $n_u=64$, $|\mathcal{I}|=8$ for NYC Taxi data. }

In conclusion, \textsf{DMPP} is moderately robust to variations in the hyper-parameters, and so can yield steady performance under different conditions.  

\subsection{Qualitative results}
To demonstrate that our model provides useful insights as to why and under which circumstances events occur,  
we analyze what was learned by our method.  

\begin{figure}[t]
\subfigure[NYC Taxi]{\centering\includegraphics[height=1.27in]{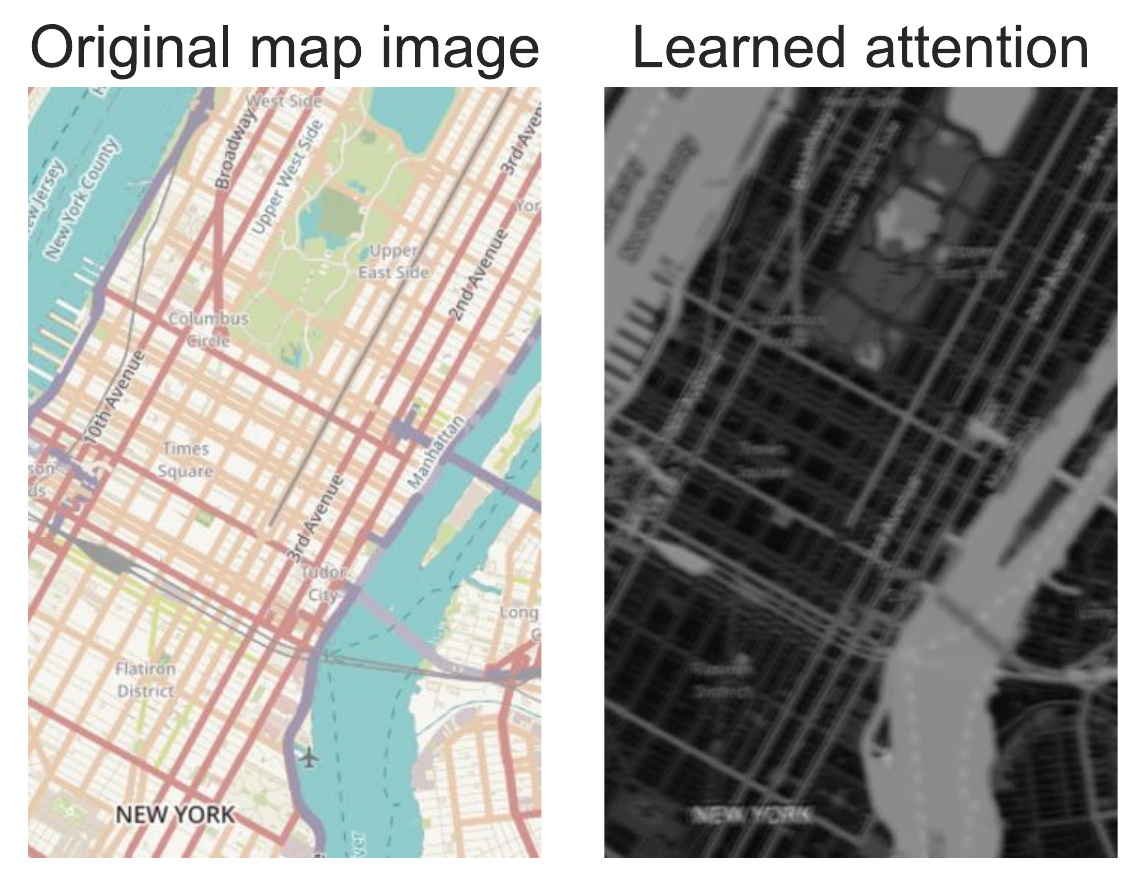}}
\subfigure[Chicago Crime]{\centering\includegraphics[height=1.27in]{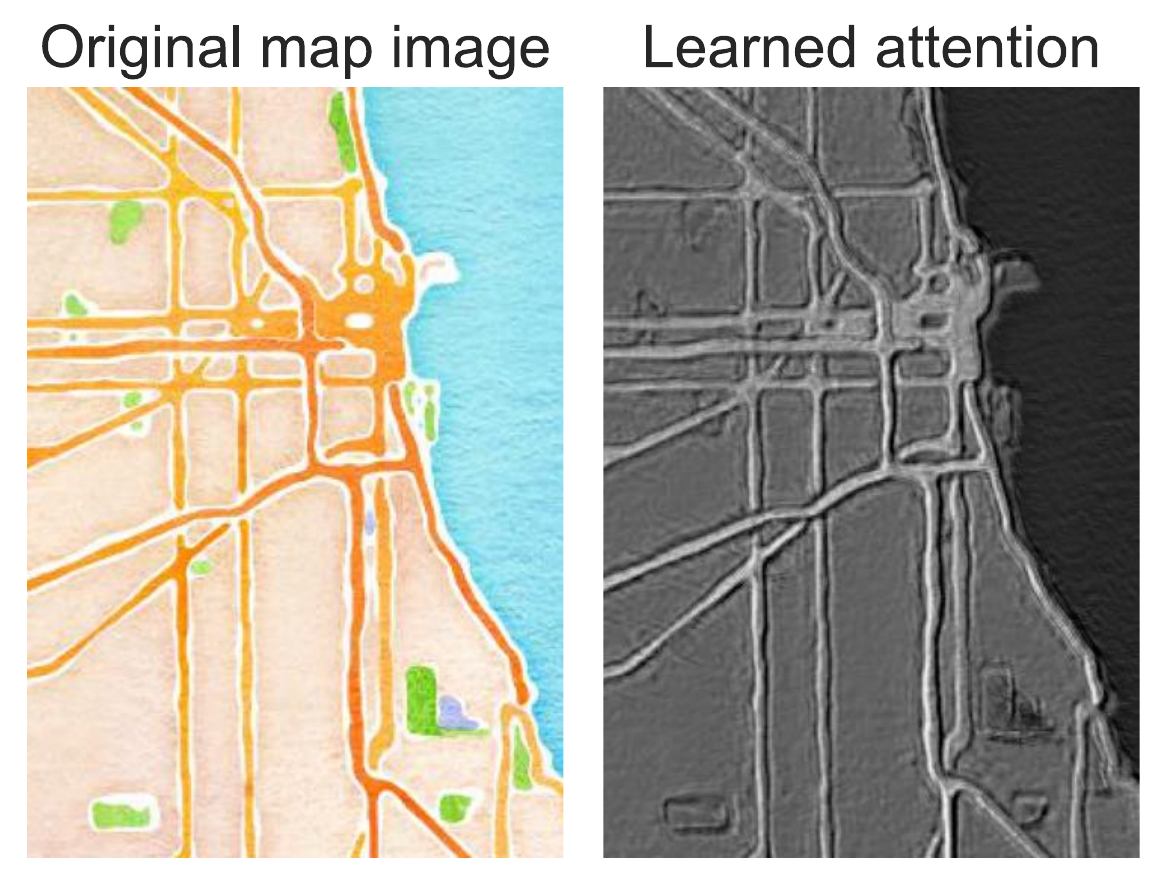}}
\caption{
Attention weights for the map images learned from NYC Taxi data and Chicago Crime. 
Left: original map image. Right: learned attention weights; 
lighter shade indicates stronger attention values. }
\label{fig:attention_map}
\end{figure}

Figure \ref{fig:attention_map} visualizes the learned attention for the map images from 
NYC Taxi data (Figure \ref{fig:attention_map}a) and Chicago Crime data (Figure \ref{fig:attention_map}b).  
Here we fed the map images (left) into the learned {\sl image attention network}, and plot the output attention weights (right). 
In the attention heatmaps (right), the light and dark regions correspond to high and low attention weights, respectively.  
We can see that \textsf{DMPP} assigns high attention weights to major roads (depicted by pink in the original map image) 
for the NYC Taxi data in Figure \ref{fig:attention_map}a. 
The minor roads (depicted by light orange) draw less attention.  
This indicates that taxi pick-ups take place mostly on roads, especially on major roads.  
Interestingly, for Chicago Crime data (Figure \ref{fig:attention_map}b), the attention mechanism weights both the roads and land cover. 
Apparently, the roads have the highest attention weights.   
This may because crime is found both on roads and in building. 
These results suggest that our method can elucidate the key spatial components related to event occurrence.  
 
\begin{figure}[t]
\centering
\hspace{-3mm}\subfigure[NYC Taxi]{\centering\includegraphics[height=1.21in]{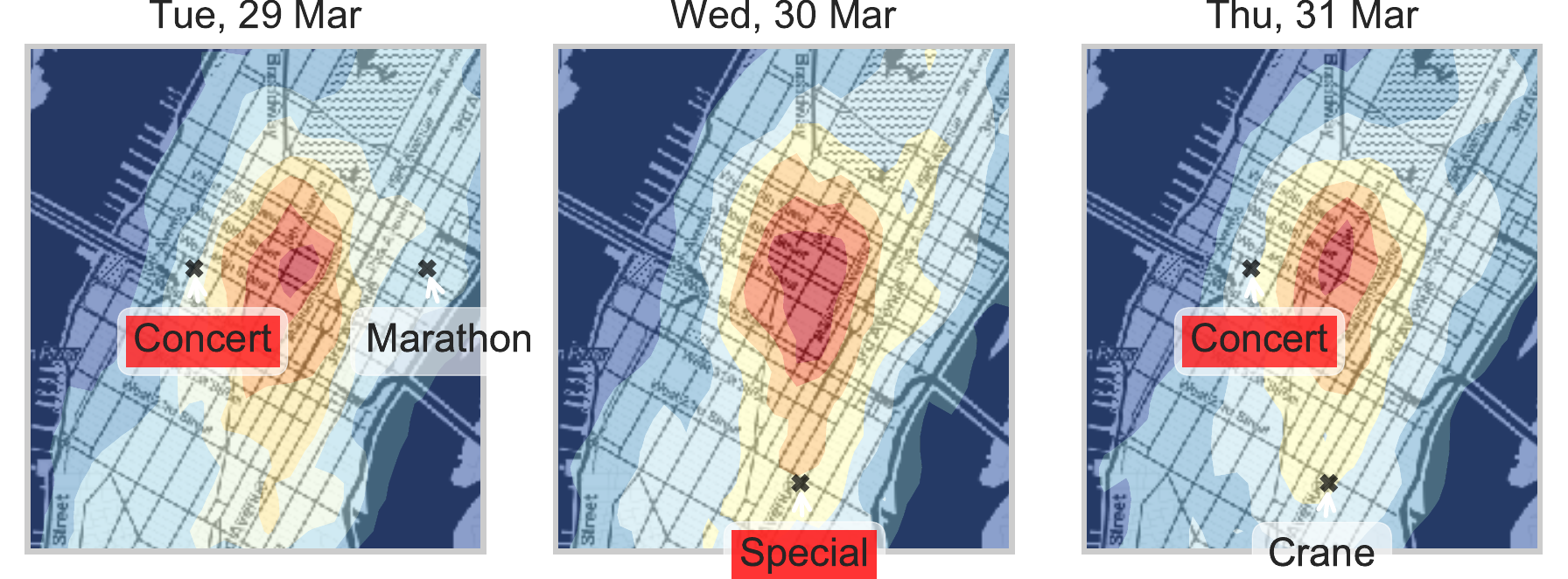}}
\subfigure[NYC Collision]{\centering\includegraphics[height=1.26in]{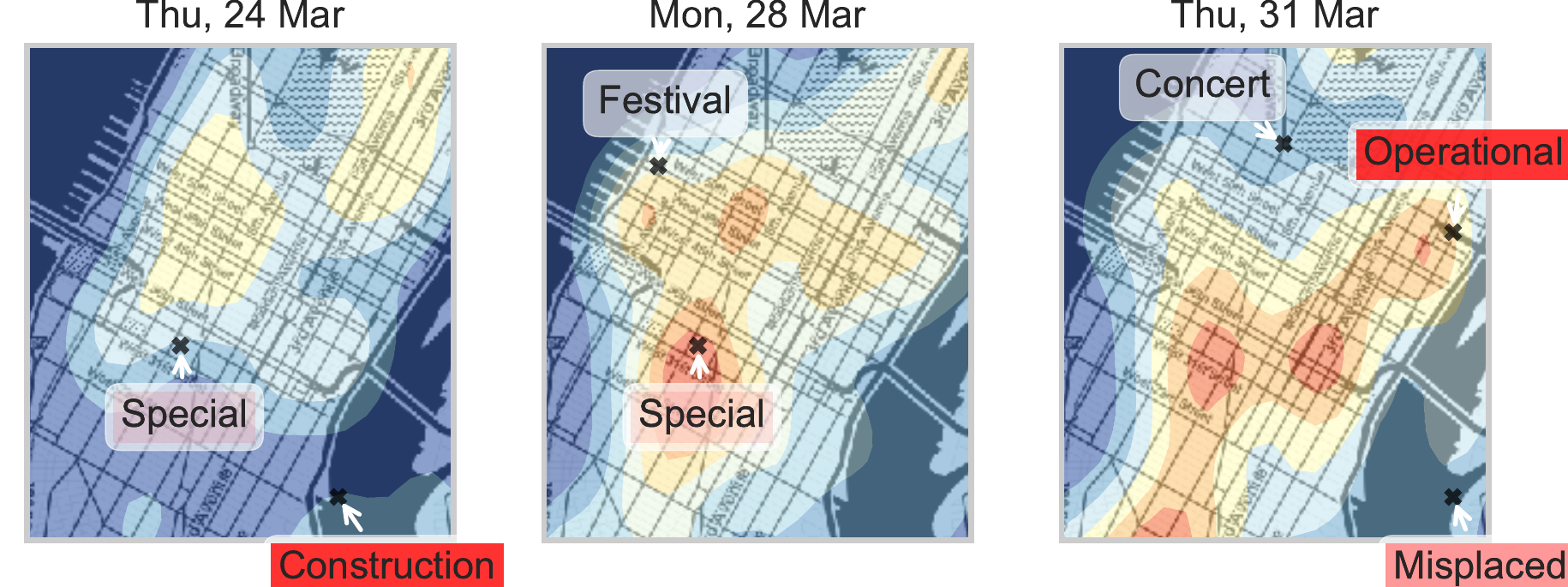}}
\caption{
Learned attention weights for social/traffic descriptions with the learned intensity around Midtown Manhattan from Mar 28th to Apr 4th.  
\textcolor{black}{Darker shade of red for the texts denotes higher attention weight.} 
}\label{fig:att_text}
\end{figure}

\begin{figure}[t]
\centering
\subfigure[NYC Taxi]{\centering\includegraphics[height=1.3in]{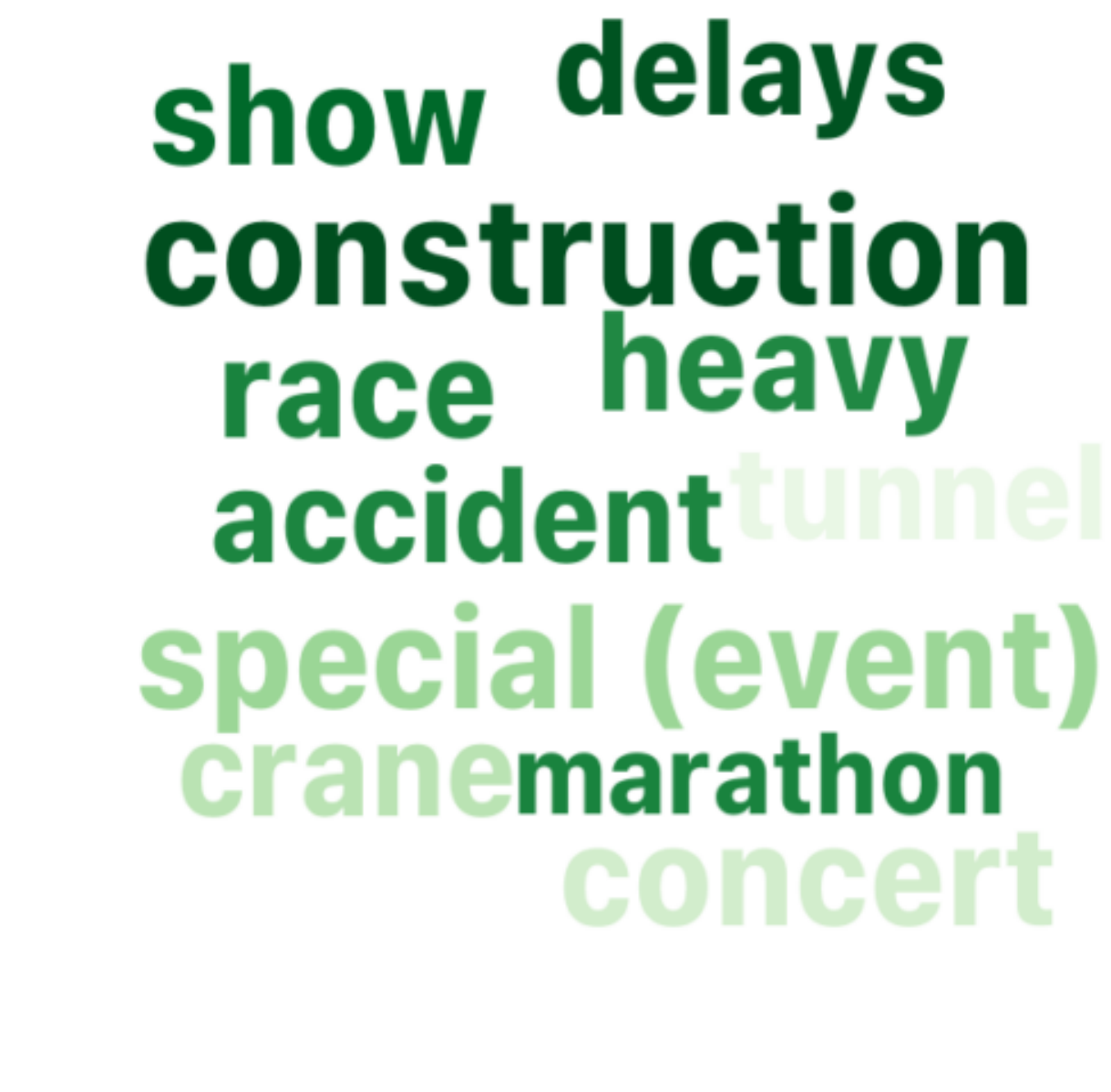}} \hspace{0.8cm}
\subfigure[NYC Collision]{\centering\includegraphics[height=1.3in]{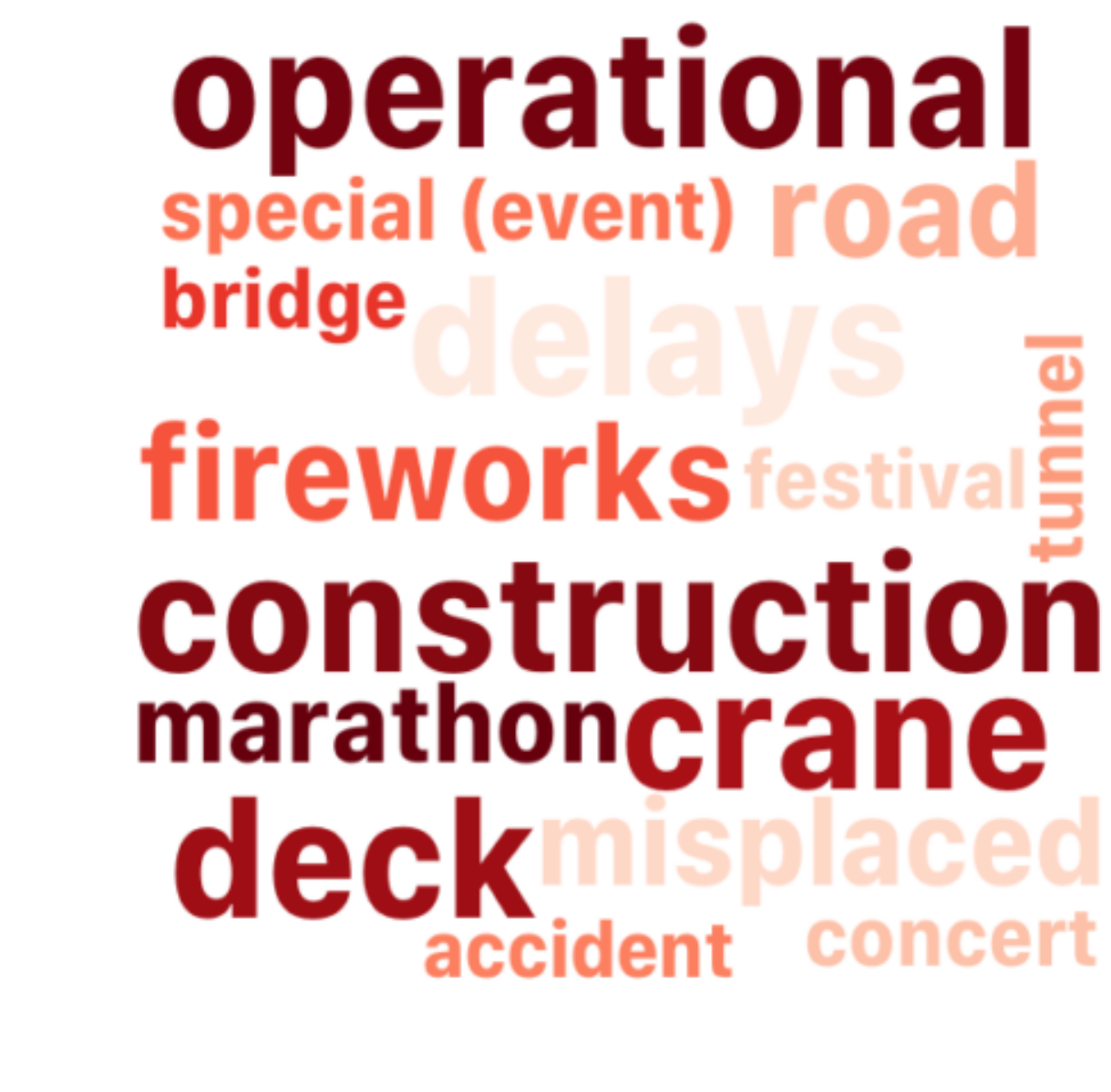}} 
\caption{Word cloud of top 15 words by attention weight;  
larger size denotes higher attention.  
}\label{fig:wordcloud}
\end{figure}

Figure \ref{fig:att_text} shows the attention weights for the event descriptions
overlaid with the learned intensity around Midtown Manhattan from Mar 24th to 31th. 
For the sake of clarity, only the first word of each sentence is depicted.  
The word {\sl Special} appears usually with {\sl event}; {\sl Operational} stands for {\sl Operational (activity)}. 
The darker shade of red for the texts indicates higher attention values.  
In the heatmaps, red corresponds to high intensity value, while blue represents low value.  
For NYC Taxi data (Figure \ref{fig:att_text}a), 
the learned intensity yields high values in Midtown and Murrey Hill of Manhattan. Seemingly, the attention mechanism also highlights the words associated with these regions. 
Mainly words related to the social events, e.g., {\sl Concert} and {\sl Special (event)}, are highlighted. 
In contrast, the words associated with traffic events, such as {\sl Construction} and {\sl Operational (activity)}, gain more attention, 
in the NYC Collision data (Figure \ref{fig:att_text}b).  
The above results suggest that traffic events affect the collision rate, whereas  social events drive the taxi pick-up demand. 
Figure \ref{fig:wordcloud} further supports this.  
It visualizes the top 15 words ranked by attention weight learned from each data set; larger size denotes higher attention.  
For NYC Taxi data, social events (e.g., {\sl special (event)} and {\sl concert}), as well as traffic event (e.g., {\sl construction}), tend to receive attnetion.  
For NYC Collision data, traffic events, e.g., {\sl construction} and {\sl operational (activity)}, seem to draw more attention.  
These results demonstrate that \textsf{DMPP} identifies important words that affect event occurrence.  
The descriptions thus found help us explain why and in which contexts events occur.  


\section{Conclusion and Future work}
In this paper, we studied the problem of event prediction in urban areas.  
Our solution, \textsf{DMPP} (Deep Mixture Point Process), is a novel point process model based on a deep learning approach.   
\textsf{DMPP} models the point process intensity by a deep mixture of kernels.  
The key advantage of \textsf{DMPP} over existing methods  
is that it can utilize the highly-dimensional and multi-sourced data provided by rich urban contexts, including images and sentences,
and automatically discover their complex effects on event occurrence.
Moreover, by taking advantage of the mixture model-based approach, 
we have developed an effective learning algorithm. 
Using real-world data sets from three different domains,  
we demonstrated that the proposed method outperforms existing methods in terms of prediction accuracy.

\bibliographystyle{ACM-Reference-Format}
\balance
\bibliography{KDD2019} 

\clearpage
\appendix
\section*{Appendix}

\section{Neural Network Architecture}
\begin{figure}[t]
\centering
\includegraphics[height=2.5in]{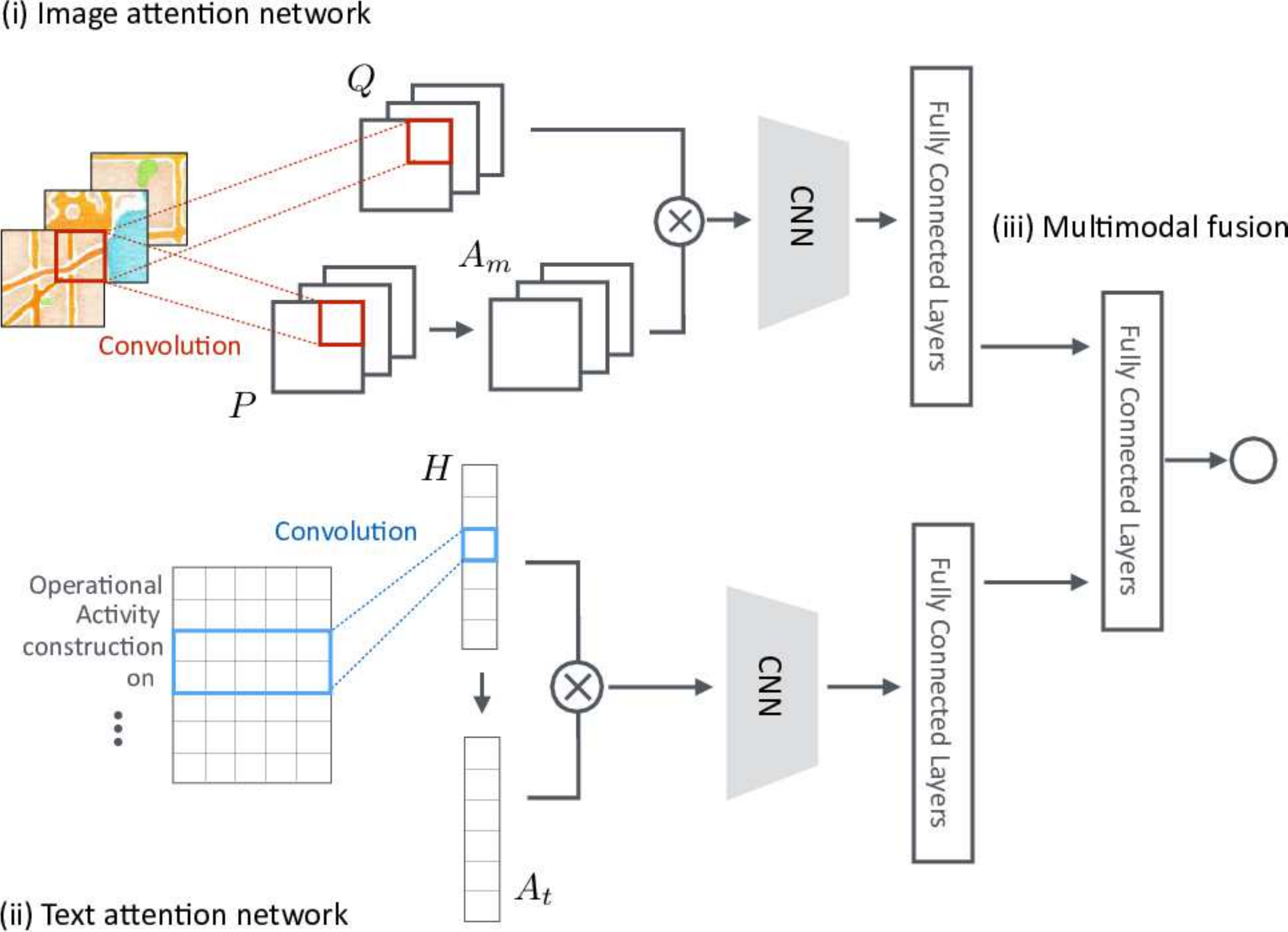}
\caption{
The architecture of the neural network used in the proposed method. 
}\label{fig:nn}
\end{figure}

This section details the architecture of the neural network used in our experiment, see Figure \ref{fig:nn}.  
Our neural network consists of three components: 
\textcolor{black}{(i)} { \sl the image attention network}, \textcolor{black}{(ii)} {\sl the text attention network}, \textcolor{black}{(iii)} {\sl the multimodal fusion module}. 
This section describes each component in detail. 
In this paper, we describe the proposal assuming the use of two types of features: map images and social/traffic event descriptions.  
Note that the proposed method can be easily extended to handle other types of features.  

\textbf{(i) Image attention network. } 
We construct the {\sl image network} by combining CNN with a self-attention mechanism, which extracts attention for regions of the image. 
Suppose we have a collection of map images 
$\{I_j\}_{j=1}^J$, $I_j\in \mathbb{R}^{N_{\text{w}}\times N_{\text{h}}\times N_{\text{c}}}$, 
where $N_{\text{w}}$, $N_{\text{h}}$, $N_{\text{c}}$ represent width, height, and the number of image features (e.g., three color channels), respectively.  
In the following discussion, we omit index $j$ for the sake of simplicity.  
{\sl The image attention network} accepts images ${I}$, and passes them through convolutional transformation followed by pooling and activation layers. 
\begin{align}\label{eq:conv}
P = g_p( C_p \ast I ),  \,\, Q = g_q( C_q \ast I )
\end{align}
where $\ast$ denotes the convolution; $C_p$ and $C_q$ are the parameter matrices to be learnt;  
$g_p(\cdot)$ and $g_q(\cdot)$ are a set of activation and pooling operations. 
\textcolor{black}{For our experiment, we use 3 $\times$ 3 same convolution so as to straightforwardly visualize the attention weights developed for the image features.} 
Subsequently, we process $P$ through a spatial attention model consisting of a single self-attention layer followed by a softmax function: 
\begin{align}
A_m = \text{softmax}\big(M_2 \text{tanh}(M_1 P^{\top})\big), 
\end{align}
where $M_1\in\mathbb{R}^{d\times d_a}$ and $M_1\in\mathbb{R}^{r\times d_a}$ are parameter matrices. 
In the experiment, we set $d=N_c$, $d_a=32, r=1$.  
The attention weights $A_m\in\mathbb{R}^{N_{\text{h}}\times N_{\text{w}}}$ indicate which regions of the image were focused on during training.  
Then, we multiply the intermediate map $P$ by the attention $A_m$. 
The output of the self-attention layer, $B$, is processed by a three layer CNN with a set of 3 $\times$ 3 convolutions. 
The output is then processed by two fully connected layers, with size of 512, and the rectified linear unit (ReLU) activation functions.

\textbf{(ii) Text attention network.}  
Social/traffic event descriptions are represented as a sequence of words $\{W_j\}_{j=1}^J$, $W_j\in\mathbb{R}^{N_{\text{s}}\times N_{\text{v}}}$,  
where $N_{\text{s}}$ is the length of the sentence and $N_{\text{v}}$ is the vocabulary size. 
We design the {\sl text network} on the CNN designed for sentences \cite{kim2014convolutional} and an attention mechanism \cite{Lin2017}.  
First, {\sl the text attention network} reads the input sequence of 1-of-K word vectors $W=[{\bf w}_{1},...,{\bf w}_{N_{\text{s}}}]$, 
${\bf w}\in\{0,1\}^{N_{\text{v}}}$ 
and transforms it into a set of hidden vectors $H=[{\bf h}_1,...,{\bf h}_{N_{\text{s}}}]$, where ${\bf h}_i$ is a $r$-dimensional vector.  
We then feed the vectors into the attention network. 
In particular, we transform the set of hidden vectors $H=[{\bf h}_1,...,{\bf h}_{N_{\text{s}}}]$ into new vectors with dimension $d_c$ such that 
\begin{align}
A_t = \text{softmax}\big(T_2 \text{tanh}(T_1 H^{\top})\big), 
\end{align}
where $T_1\in\mathbb{R}^{r}$ and $T_1\in\mathbb{R}^{r}$ are parameter matrices.
\textcolor{black}{We set $r=1$ in our experiment. }
$A_t\in\mathbb{R}^{N_{\text{s}}}$, the attention weight for each word, reflects the importance of the word.  
We multiply the hidden vectors $H$ with the attention weights $A_t$, and feed the results through a three layer CNN.  
The output of the CNN is transformed by two fully connected layers with size of 8, ReLU activation functions, and dropout of 0.1.  

\textbf{(iii) Multimodal fusion module.} 
The positions of the representative points, {\bf u}, are processed by two fully connected layers with 32 units and relu activations. 
Their output and the outputs of the attention network ({\sl the image attention network} or {\sl text attention network}) are concatenated.  
This is followed by fully connected layers with relu activation functions. 
The hyper-parameters of the last fully connected layers for {\sl the multimodal fusion module} are tuned on the validation set.  

\section{Implementation}
\subsection{Kernel function}
As mentioned in 5.2, we explored three kinds of kernel functions: uniform, Gaussian, and compactly supported Gaussian.  
We define each kernel below.   

{\bf Uniform kernel. }
\begin{align}
k({\bf x},{\bf u}_j) = \mathbbm{1}{ (||{\bf x}-{\bf u}_j|| < w) }, \nonumber
\end{align}
where $\mathbbm{1}{(\cdot)}$ is an indicator function. 

{\bf Gaussian kernel. }
\begin{align}
k({\bf x},{\bf u}_j) = \exp{\big(- ({\bf x}-{\bf u}_j)^\top \Sigma^{-1} ({\bf x}-{\bf u}_j)\big)}, \nonumber
\end{align}
where $\Sigma$ is a $3\times 3$ covariance matrix.  

{\bf Compactly supported Gaussian kernel. }
\begin{align}
k({\bf x},{\bf u}_j) = \exp{\big(- ({\bf x}-{\bf u}_j)^\top \Sigma^{-1} ({\bf x}-{\bf u}_j)\big)} \cdot
\mathbbm{1}{ (||{\bf x}-{\bf u}_j|| < w) }, \nonumber
\end{align}
where $\Sigma$ is a $3\times 3$ covariance matrix, $\mathbbm{1}{(\cdot)}$ is an indicator function, and 
$w$ is a positive parameter that thresholds the kernels, $||{\bf x}-{\bf u}_j|| \geq w$, to zeros.  
This means that $k({\bf x}, {\bf u}_j)$ will be zero when ${\bf x}$ and ${\bf u}_j$ are far enough away.  
The use of the compactly supported kernel allows for an effective learning algorithm, especially for large data size $N$ and 
for large numbers of representative points, $J$.  
The objective (\ref{eq:like}) involves kernel evaluations for all pairs of ${\bf x}_i$ and ${\bf u}_j$,  
resulting in an $N\times J$ kernel matrix $K$ with elements $K_{ij}=k({\bf x}_i,{\bf u}_j)$. 
The back-propagation is carried out by taking the derivation of $K$, 
which requires $\mathcal{O}(NJ)$ operations at each iteration ($\mathcal{O}(|\mathcal{I}| J)$ for the mini-batch optimization).  
The computation burden can be impractically heavy when the data size $N$ (the mini-batch size $|\mathcal{I}|$) or the number of representative points $J$ is large. 
The use of the compactly supported kernel allows us to scale up the learning algorithm. 
Such a kernel ensures that the kernel element $K_{ij}$ is zero whenever the distance between ${\bf x}_i$ and ${\bf u}_j$ is above a certain threshold.  
This leads to a sparse structure in the kernel matrix, $K$, thus allowing for fast sparse matrix computations.  

\subsection{Parameters}
For each representative point, we extract a 20 $\times$ 20 image patch from the map image of the entire region-of-interest (i.e., Manhattan for NYC Collision data and NYC Taxi data, City of Chicago for Chicago Crime data), resize it to 10 $\times$ 10 pixels, and use its RGB vector as the input image.  
This corresponds to 290m$\times$ 500m square grid space for NYC Collision data and NYC Taxi data, 
and 290m$\times$ 600m square grid space for Chicago Crime data.  
Therefore, $N_w=10$, $N_h=10$, $N_c=3$.  
In {\sl the text network}, we only consider the first 200 most frequent words, and use the first 5 words of each sentence. 
For our input descriptions, we zero-pad to ensure a sentence length of 5 words. 
Thus, $N_v=200$ and $N_s=5$. 

\subsection{Environment}
\textsf{RMTPP} and our \textsf{DMPP} are implemented using the Chainer deep network toolkit \cite{tokui2015chainer}. 
All the methods are run on a Linux server with an Intel Xeon CPU, and a GeForce GTX TITAN GPU. The GPU code is implemented using CUDA \textcolor{black}{9}.  

\section{Compared methods}
In Section 5, we compared the proposed method with the existing methods in terms of the predictive log-likelihood per event (LogLike). 
Below we describe the setting and the procedure adopted for each method in detail.   

{\bf HP (Homogeneous Poisson process).}
The intensity of \textsf{HP} is assumed to be constant over space and time: $\lambda({\bf x})=\lambda_0$.  
Given the test period $[T,T+\Delta T]$ and the region of interest $\mathbb{S}$, 
the likelihood of \textsf{HP} is written as 
\begin{align}
\log{p(\mathcal{X}|\lambda({\bf x})=\lambda_0)} &= n \log \lambda_0 - \lambda_0 \Delta T |\mathbb{S}|, 
\end{align}
where $n$ is the number of test samples, $\Delta T$ is the length of the test period, and $|\cdot|$ is the operator providing the area of a spatial region.

{\bf LGCP (Log Gaussian Cox process). } 
We implement the general LGCP method described in \cite{Panik2010}, where the intensity is defined as  
$\lambda({\bf x}) = \mu(t) \psi(s) \exp{\big(y({\bf x})\big)}$; 
$\mu(t)$ and $\psi(s)$ are temporal and spatial background rates, respectively.  
$y(\cdot)$ is a Gaussian process with the following covariate function:  
\begin{align}
\text{cov}({\bf x},{\bf x}') = \sigma_{\text{GP}}^2 \exp{(-({\bf x}-{\bf {x}'})^{\top}\theta_{\text{GP}}({\bf x}-{\bf {x}'}))}, 
\end{align}
\textcolor{black}{where $\sigma_{\text{GP}}$ is the scale parameter, $\theta_{\text{GP}}$ is the bandwidth. }

{\bf RMTPP (Recurrent Marked Temporal Point Process).}
As for \textsf{DMPP}, we used the Adam algorithm \cite{Kingma2014} 
with $\beta_1 = 0.01$, $\beta_2 = 0.9$ and learning rate of 0.01 for \textsf{RMTPP}.  
Following \cite{Du2016}, we map the coordinates using NYC Neighborhood Names GIS dataset\footnote{https://data.cityofnewyork.us/City-Government/Neighborhood-Names-GIS/99bc-9p23} for NYC Collision and NYC Taxi data.  
Similarly, we use Chicago Neighborhood Names GIS dataset\footnote{https://data.cityofchicago.org/Facilities-Geographic-Boundaries/Boundaries-Neighborhoods/bbvz-uum9} for Chicago Crime data.  
Finally, we obtained 48 unique locations for NYC Collision and NYC Taxi data, 44 for Chicago Crime data. 
We set unit size as 16 and batch size as 32 for NYC Collision data,  
unit size as 16 and batch size as 8 for Chicago Crime data,  
unit size as 16 and batch size as 16 for NYC Taxi data.  
We use the log-likelihood (LogLike) as defined in \cite{Du2016}.  
\textcolor{black}{
For event number prediction, we generate the sequence of events by sequentially predicting the timing of the next event.  
In particular, given the event sequence $\{t_1,...,t_N\}$, we compute the timing of next event $\hat{t}_{N+1}$, using Equation (13) in \cite{Du2016}. 
Using $\hat{t}_{N+1}$ as known data, we then predict the timing of next event $\hat{t}_{N+2}$ based on the new sequence $\{t_1,...,t_N,\hat{t}_{N+1}\}$.  
This procedure is repeated until $\hat{t}_{N+i}>T+\Delta T$. 
}

\end{document}